%% file: acl_latex.tex
\documentclass[11pt]{article}

\usepackage[final]{acl}

\usepackage{times}
\usepackage{latexsym}

\usepackage[T1]{fontenc}

\usepackage[utf8]{inputenc}

\usepackage{microtype}

\usepackage{inconsolata}

\usepackage{comment}
\usepackage{xspace}
\usepackage{multirow}
\usepackage{multicol}
\usepackage{graphicx}
\usepackage{booktabs}
\usepackage{todonotes}
\usepackage{tablefootnote}
\usepackage{soul}
\usepackage{subfigure}
\usepackage{listings}
\usepackage{pgfplots}
\usepackage{amssymb}
\usepackage{amsmath}
\usepackage{soul,xcolor}
\usepackage{inconsolata}
\usepackage[normalem]{ulem}
\usepackage[most]{tcolorbox}
\newcommand{\method}{\textsc{AnalysisBank}\xspace}
\definecolor{condBlue}{HTML}{D6E8F5}
\definecolor{condAmber}{HTML}{F8E5C2}
\definecolor{condGreen}{HTML}{D8ECDE}
\definecolor{condPurple}{HTML}{E4D9F0}
\definecolor{amber}{HTML}{D89B2C}

\usepackage{xcolor, soul,todonotes} 
\definecolor{kmycolor}{rgb}{0.858, 0.188, 0.478}

\newcommand{\hlBlue}[1]{\sethlcolor{condBlue}\hl{#1}}
\newcommand{\hlAmber}[1]{\sethlcolor{condAmber}\hl{#1}}
\newcommand{\hlGreen}[1]{\sethlcolor{condGreen}\hl{#1}}
\newcommand{\hlPurple}[1]{\sethlcolor{condPurple}\hl{#1}}
\title{\method: An Expert Analysis Pattern Library for \\ Financial Report Generation}

\author{
    Yajing Yang$^{1}$, Yunshan Ma$^{2}$, Kelvin J.L. Koa$^{1,3}$, Min-Yen Kan$^{1}$
    \vspace{2mm} \\
    \textsuperscript{1}National University of Singapore\;\;  \textsuperscript{2}Singapore Management University\\ \textsuperscript{3}Asian Institute of Digital Finance\\
    {\texttt{\small yajing.yang@u.nus.edu\;;
ysma@smu.edu.sg\;; kelvin.koa@u.nus.edu\;; kanmy@comp.nus.edu.sg}}
}

\begin{document}
\maketitle
\begin{abstract}

We argue that financial report generation should operate at the analytical rather than structural level, composing content from data-derived insights rather than high-level topics or sections. To this end, we propose \method, which distills expert reports into a reusable library of \emph{Analyses}, each pairing a data signal, an analytical move, and the expert span it was derived from. At inference time, \method matches input signals to library entries and applies the retrieved moves to compose the report. A study of \emph{Analyses} distilled from 550 expert reports reveals a heavy-tailed distribution of 47--52 signal types spanning 13 move types. On two financial benchmarks across four LLM backbones, \method increases the proportion of novel, data-grounded insights by 1.7--3.7$\times$ over structural-level baselines. Transfer to scientific writing suggests that the distinction generalizes beyond finance. Code and the distilled \emph{Analysis} library are available at \url{https://github.com/yajingyang/AnalysisBank}.
\end{abstract}

\section{Introduction}
\label{sec:intro}


In finance, analytical reports support decisions by producing \emph{insights}: claims derived from analyzing source data rather than paraphrasing it~\cite{asquith2005information, huang2014evidence}. Each insight is the output of an \emph{analytical move}: a comparison, an attribution, a derivation, a projection, a gap-detection. Which analytical  moves to make depends on what the data contains: the same earnings transcript, read by different analysts, yields different insights because each analyst recognizes different conditions in the data and responds with different analytical moves~\cite{huang2018analyst}. This distinguishes the task from summarization~\citep{lewis2020bart} and data-to-text generation~\citep{wiseman2017challenges}, where the output paraphrases or restates the input; analytical report generation produces content the input does not state.

\begin{figure}[tb]
    \centering
    \includegraphics[width=1\columnwidth]{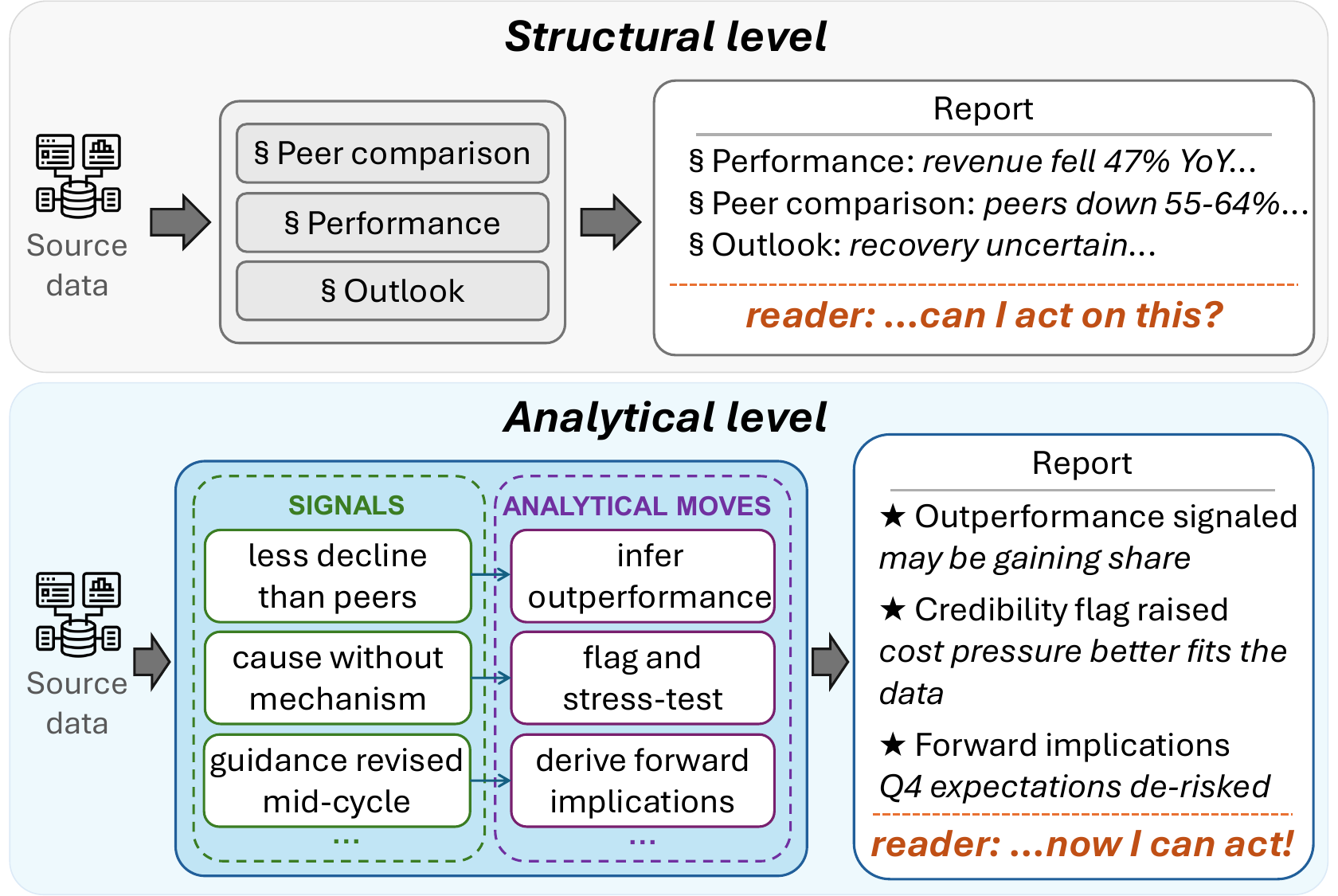}
    \caption{\textbf{Structural vs.\ analytical report generation.} Structural level (top): methods prescribe sections but leave the reasoning to model defaults. Analytical level (bottom): data signals are matched to the analytical moves they call for, yielding data-driven insights.}
    \label{fig:task}
     \vspace{-0.3cm}  
\end{figure}

Existing methods operate at what we term the \emph{structural level}: they prescribe what the report should cover, such as topics, viewpoints, or analytical categories, but leave the reasoning within each section to the language model's default behavior~\cite{goldsack2025facts, koshkin2025margen, yang2022re3, shao2024storm} (Figure~\ref{fig:task}, top). In every case the skeleton varies but the analytical substance does not: sections are filled with generic analyses such as trend description and standard ratio computation, regardless of what the input data specifically warrants.
We argue that the task requires operating at the \emph{analytical level}: the report should be composed of analytical moves driven by the specific signals in the source data (Figure~\ref{fig:task}, bottom). For example, a revenue decline smaller  than peers' calls for a relative-outperformance inference; a downward guidance revision calls for a margin-implication derivation. These are not generic trend descriptions but signal-specific insights.

Operating at the analytical level is non-trivial for two reasons. First, the space of pairings between data signals and the analytical moves they call for is large and heavy-tailed: our corpus analysis identifies 47--52 signal types spanning 13 analytical move types, yet over a third appear in three or fewer instances, each pairing with only one to two moves (\S\ref{sec:bank-analysis}). These tail patterns 
reflect the diverse ways experts respond to uncommon data conditions~\citep{huang2018analyst} and are too sparse for LLMs to learn from prompting or fine-tuning alone.
Second, even given comprehensive pairings, selecting the right one for a given input is challenging: 
as autoregressive models, LLMs are biased toward high-probability outputs~\citep{mccoy2024embers} and default to generic analytical moves rather than the rare, signal-specific move the data calls for. 
Our ablation confirms that alternate selection mechanisms yield 19\% less insightful content (\S\ref{sec:design-space}).

To close this gap, we propose \method. The core idea is that expert reports already embody analytical-level reasoning: each insight is derived from a specific data signal with the analytical move it calls for. \method distills these pairings into a reusable library, where each entry, an \emph{Analysis}, tuples a data signal, an analytical move, and the expert span it was distilled from, with the signal and move abstracted to making each \emph{Analysis} reusable across inputs sharing the same pattern. Our study shows that \emph{Analyses} distilled from 550 expert reports cover the heavy-tailed distribution (\S\ref{sec:bank-analysis}). 
The same abstraction addresses the selection challenge: by disentangling the underlying analytical pattern from surface specifics, the data signal field makes the link between a stored pattern and the input signal it should match explicit, enabling direct retrieval at inference time.
On two benchmarks covering structured market data and unstructured earnings transcripts, \method{} increases the proportion of novel, data-grounded insights by 1.7--3.7$\times$ across four LLM backbones.

Our contributions are threefold: (i) we distinguish \emph{structural} and \emph{analytical} levels of report generation and argue existing methods address only the former; (ii) we propose \method, a library of expert-distilled \emph{Analyses}  and a pipeline to apply them at inference; (iii) on two benchmarks across four backbones, \method increases novel, data-grounded insight by 1.7--3.7$\times$, while structural-level methods converge well below it.

\section{Related Work}
\label{sec:related-work}
We break down relevant prior work into three areas: long-form and report generation (the task), reusable reasoning patterns (the method), and financial NLP (the domain).

\paragraph{Long-form and report generation.} Data-to-text generation maps tables and records to fluent prose~\cite{wiseman2017challenges,parikh2020totto}, but  the output restates the input rather than analyzing it.
Report-generation methods go further by introducing organizational structure:
multi-agent frameworks assign predefined analytical roles~\cite{goldsack2025facts,koshkin2025margen}, 
plan-then-write approaches decompose output into sections~\cite{yang2022re3,puduppully2019content,hu2022planet},
and retrieve-organize-write systems research a topic through retrieval and generate from the resulting outline~\cite{shao2024storm}.
All three control what the report covers but not what analysis to apply to the data.

\paragraph{Reusable reasoning patterns.} 
A natural approach is to retrieve reusable reasoning patterns at inference time. Self-discovered reasoning modules~\cite{zhou2024selfdiscover} and thought templates~\cite{yang2024bot,jeong2025total} provide abstract reasoning strategies not grounded in specific data patterns. Agent workflows~\cite{wang2024awm,wang2024voyager} and induced functions~\cite{wang2024trove} learn concrete routines from task-solving trajectories but capture full procedures, not individual analytical moves. Case-based methods~\cite{yasunaga2024analogical,wiratunga2024cbrrag} and experience memories~\cite{park2023generative,zhong2024memorybank} operate at the instance level but store complete cases tied to specific inputs, not abstracted into reusable patterns. Analytical report generation requires patterns that are grounded in data, decomposed at the analytical-move level, and abstracted for reuse across the heavy-tailed distribution found in expert reports.

\paragraph{Financial NLP.} Domain-pretrained models~\cite{wu2023bloomberggpt,yang2023fingpt, xie2023pixiu} adapt language
models to financial text but do not explicitly model
signal-to-move reasoning.  Signal-mining pipelines extract  salient content within it~\cite{ju2023signals,lu2025extraction} but do not prescribe what analysis a signal calls for. Downstream reasoning work targets question answering~\cite{chen2021finqa,zhu2021tatqa,islam2023financebench,finder2025choi}, injecting expert structure through hand-authored workflows~\citep{nitarach2025fincot} or historical  scaffolds~\citep{adfcot2025, augmentedllm2024}, but produces closed-form answers, not open-ended reports. For report generation, bullet-point earnings-call summarization~\cite{mukherjee2022ectsum}  remains structural, while hierarchical data narration~\citep{yang2025kahan} reaches the analytical level but leaves the choice of analysis to the model, which defaults to the most common ones.

\section{\method}
\label{sec:method}

We first define the representation of each \emph{Analysis}, designed to be reusable across inputs while preserving actionability (\S\ref{sec:representation}), then the extraction pipeline that populates the library from a corpus of expert reports (\S\ref{sec:extraction}), and finally characterize the resulting library (\S\ref{sec:bank-analysis}).

\subsection{\emph{Analysis} Representation}
\label{sec:representation}

\begin{table}[t]
\centering
\small
\renewcommand{\arraystretch}{1.3}
\begin{tabular}{@{}p{0.98\linewidth}@{}}
\toprule
\textbf{\texttt{data\_signal}}:
A key revenue or fee line shows \hlBlue{a large decline versus the prior period}, but the decline is \hlAmber{smaller than what was broadly expected} and also \hlGreen{smaller than the decline reported by a comparable peer} facing similar conditions. \\
\midrule
\textbf{\texttt{analytical\_move}}:
When a \hlBlue{major revenue or fee category declines sharply}, compare the actual change to \hlAmber{pre-report expectations} and to \hlGreen{the change reported by close peers}; if the decline is less severe than both, \hlPurple{explicitly assess and quantify whether this indicates relative outperformance and potential share gains} in that activity. \\
\midrule
\textbf{\texttt{reference\_text}}:
``Even investment banking, widely expected to be a sore thumb for all banks this quarter, \hlPurple{may have contributed to this Friday's bullish reaction}. To be clear, revenues of \$1.7 billion were small as \hlBlue{fees sank by a whopping 47\% YOY}. But keep the following in mind: (1) \hlAmber{expectations were highly de-risked ahead of the print, with Seeking Alpha contributor Cavenagh Research bracing for a fee drop greater than 50\%}, and (2) JPMorgan, the top Wall Street player in investment banking, \hlPurple{may have stolen some market share away from its competitors once again}. \hlGreen{Citi (C), for example, reported Q3 investment banking revenues that were an astonishing 64\% lower YOY}.'' \\
\bottomrule
\end{tabular}
\caption{Example \emph{Analysis}. The data signal with three co-occurring conditions (\hlBlue{actual vs. prior period}, \hlAmber{actual vs. expectations}, \hlGreen{actual vs. peer}) feeds one analytical move (\hlPurple{inference of outperformance and share gains}).}
\label{tab:thought-example}
\vspace{-0.3cm}
\end{table}

\begin{figure*}
    \centering
    \includegraphics[width=1\textwidth]{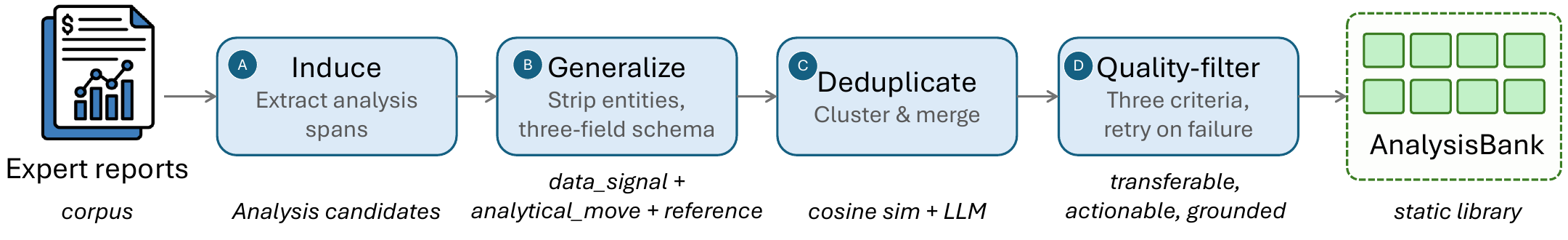}
    \caption{The four-pass \method extraction pipeline, converting a corpus of expert reports into a static library of \emph{Analyses}.}
    \label{fig:extraction-pipeline}
    \vspace{-0.3cm}
\end{figure*}

Each entry in the \method{} library, an \emph{Analysis}, is a tuple of (\texttt{data\_signal}, \texttt{analytical\_move},
\texttt{reference\_text}), as illustrated in
Table~\ref{tab:thought-example}. 

The \texttt{reference\_text} is a verbatim span from the source expert report that demonstrates the analytical move applied to real data (e.g., ``revenues of \$1.7 billion were small as fees sank by a whopping 47\% YOY''). It provides a generation anchor and a faithfulness check, but contains specific entities, figures, and time references that prevent it from transferring to new inputs.

The \texttt{analytical\_move} abstracts the reasoning in the reference text into a reusable instruction specifying what analysis to perform on the matched signal (e.g., ``compare the actual change to expectations and to peers; if less severe than both, assess whether this indicates relative outperformance and potential share gains''). This transfers across inputs, but its specificity hinders fuzzy retrieval when used as the matching key.

The \texttt{data\_signal} resolves this by serving as a separate retrieval key, stating triggering conditions in entity-free, number-free language (e.g., ``a key revenue line declines sharply, but less than broadly expected and less than a comparable peer''). Retrieval matches on structural patterns while generation follows the specific instruction in the analytical move.

This three-field decomposition is a design hypothesis; we evaluate it against alternative designs in \S\ref{sec:design-space}, finding that removing or merging any field degrades either retrieval precision or generation quality.

 
 \begin{figure*}[t]
    \centering
    \includegraphics[width=\textwidth]{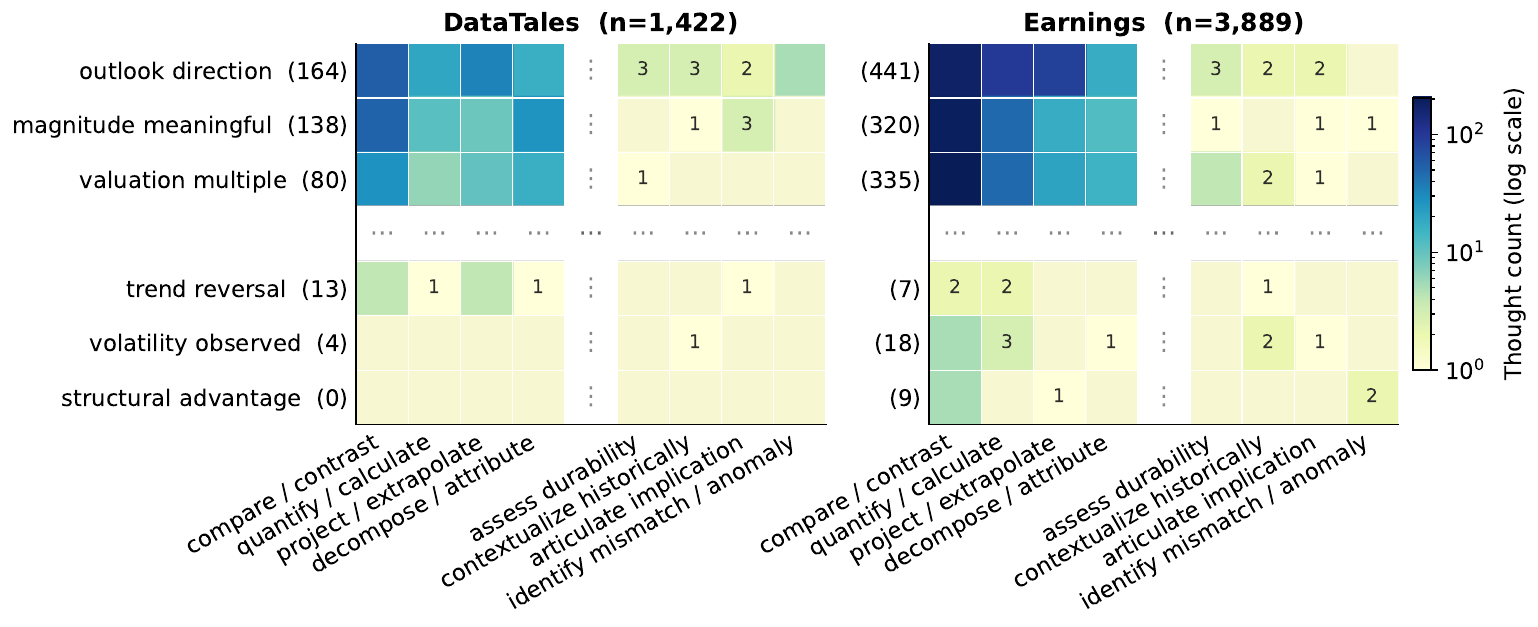}
    \vspace{-0.7cm}
    \caption{Signal-by-move distribution for three head and three tail signal types. Rows are signal types (parenthesized counts = total \emph{Analyses}); columns are analytical-move types; cell color encodes \emph{Analysis} count on a log scale. Full distribution in Appendix~\ref{app:bank-details}.}
    \label{fig:bank-heatmap}
    \vspace{-0.3cm}
\end{figure*}

\subsection{Extraction Pipeline}
\label{sec:extraction}

Prior pattern libraries induce entries from problem-solving traces with observable success signals (cf \S\ref{sec:related-work}); expert reports offer no such structure, requiring the pipeline to reverse-engineer the three fields from finished prose. We design a four-pass pipeline to convert the corpus into a static library (Figure~\ref{fig:extraction-pipeline}). 

\textbf{Induce} identifies candidate \emph{Analysis} content in each report. Analytical moves are implicit in expert prose; this stage makes them explicit by locating text spans where the analyst performed a specific reasoning operation. These spans become the \texttt{reference\_text} of each \emph{Analysis}.

\textbf{Generalize} extracts the \texttt{data\_signal} and \texttt{analytical\_move} from each identified span, producing the full three-field \emph{Analysis} of \S\ref{sec:representation}. The data signal is abstracted by stripping entity names, specific figures, and time references while preserving the conditional structure. The analytical move is similarly abstracted into a specific instruction that applies to any instance of the signal. A self-check enforces that each \emph{Analysis} would apply to the same pattern across different industries.

\textbf{Deduplicate} drops redundant \emph{Analyses} across reports, since multiple expert reports often apply the same analytical move to similar signals. Candidates with \texttt{data\_signal} embeddings above a cosine similarity threshold are grouped together and merged into a single \emph{Analysis}, retaining the most general data signal and the most complete analytical move. This prevents duplicate entries from consuming retrieval budget and biasing toward head patterns.

\textbf{Quality-filter} validates each \emph{Analysis} against three criteria aligned with the three fields: transferability (does the data signal generalize across industries?), actionability (would two analysts following the analytical move independently produce similar outputs?), and grounding (does the reference text show a real instance of the pattern?). Grounding failures are discarded; transferability or actionability failures are routed back to Generalize with a targeted rewrite instruction. The resulting \emph{Analyses} are industry-agnostic, executable, and empirically grounded.

The pipeline runs offline once, allowing a stronger model than the inference backbone to maximize \emph{Analysis} quality. The resulting library is reused across all generation runs.

\begin{figure*}[t]
    \centering
    \includegraphics[width=\textwidth]{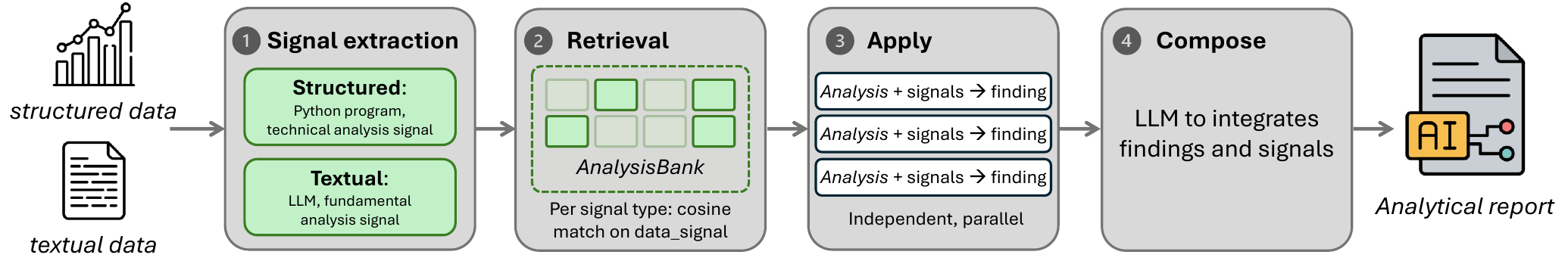}
    \caption{\textbf{Report generation pipeline.} A source input is reduced to typed signals (Stage~1), each signal type retrieves the most relevant \emph{Analysis} from \method{} library as constructed in Figure~\ref{fig:extraction-pipeline} (Stage~2), each retrieved \emph{Analysis} is independently applied to produce a finding (Stage~3), and the findings are composed into a report (Stage~4).}
    \label{fig:pipeline}
    \vspace{-0.3cm}
\end{figure*}

\subsection{Coverage and Distribution}
\label{sec:bank-analysis}


We instantiate \method on two corpora spanning the dominant input modalities: 550 daily market-analysis reports written from structured price data (\emph{DataTales}; \citealp{yang2024datatales}) and 550 equity-research reports written from filings and earnings-call transcripts (\emph{Earnings}\footnote{The Earnings \emph{Analysis} library is distilled from Seeking Alpha (\url{https://seekingalpha.com/}) analyst reports, which we cannot redistribute.}). The extraction pipeline of \S\ref{sec:extraction} yields 1{,}422  \emph{Analyses}  from DataTales and 3{,}889 from Earnings. We characterize each library by labeling every \emph{Analysis} with a signal type and a move type using two hand-built keyword taxonomies (Appendix~\ref{app:bank-details}).
The move taxonomy has 13 types (e.g., compare/contrast), all populated in both corpora. The signal taxonomy has 55 types (e.g., outlook direction), of which 47 signal types are populated in DataTales and 52 in Earnings. The two axes are not independent: each signal type triggers a median of 4 distinct move types in DataTales and 6 in Earnings, and the move mix varies across signals (e.g., outlook direction spreads across compare/contrast, assess durability, contextualize historically, and articulate implication, while volatility observed concentrates on articulate implication alone; Figure~\ref{fig:bank-heatmap}).

This breadth coexists with a heavy long tail. The top-5 signals account for only 55\% and 51\% of \emph{Analyses} in DataTales and Earnings respectively; reaching 90\% coverage requires 19 and 22 signal types. The remaining tail is not residual noise: 17 of 47 signal types in DataTales (36\%) and 10 of 52 in Earnings (19\%) appear in ${\leq}3$ \emph{Analyses}, yet unlike head signals that spread across many moves, each tail signal concentrates on 1--2 moves (Figure~\ref{fig:bank-heatmap}, below the dotted line), reflecting the narrow, decisive way an expert responds to an uncommon situation such as a trend reversal or a structural advantage assessment. These are precisely the patterns too rare for prompting or fine-tuning to internalize, preserved in reusable form by the abstraction of \S\ref{sec:representation}.

\section{Narrative Generation with \method}
\label{sec:narration-pipeline}

The library built in \S\ref{sec:method} provides coverage and the field structure needed for effective retrieval; the generation pipeline selects and applies the right \emph{Analyses} for a given input (Figure~\ref{fig:pipeline}). 

\paragraph{Stage 1: Signal extraction.}
Raw financial input is too broad to match effectively against the abstracted \texttt{data\_signal} field. This stage reduces it to a list of typed \emph{signals}, each consisting of a signal type (e.g., margin delta, volatility regime), a signal description summarizing the condition, and supporting source spans. For textual input (e.g., earnings-call transcripts), an LLM extracts the signals, each verified by a fact-checking pass~\citep{penman2013financial}. For structured input (e.g., market data), a Python program computes them deterministically from price and volume series~\citep{murphy1999technical} and formats the results into the signal description for retrieval.

\paragraph{Stage 2: Per-type retrieval.}
Signal descriptions query \method{} by cosine similarity against \texttt{data\_signal} field of each library entry. To ensure the retrieved \emph{Analyses} cover diverse signal types rather than concentrating on one, each signal type contributes its best matching \emph{Analysis}, and remaining slots are filled by global top-$k$.

\paragraph{Stage 3: Per-\emph{Analysis} execution.}
Each retrieved \emph{Analysis} is applied independently: an LLM call executes the \texttt{analytical\_move} on the triggering  signal descriptions and their supporting spans, producing one analytical finding. The \texttt{reference\_text} is included as a demonstration of how the same type of analytical move has been applied before.
Independence across calls ensures that Stage~4  receives findings with distinct analytical content. 

\paragraph{Stage 4: Composition.}
A single LLM call composes the findings into a report. Rather than following a fixed section template, the model organizes around investment-relevant themes inferred from the findings themselves, consistent with the analytical-level principle that report structure should be driven by the data.
The extracted signals  from Stage~1 are included to fill factual gaps the findings do not cover, grounding the report in the source data.\\

The pipeline uses the same backbone model at every stage. Prompts and hyperparameters are in Appendix~\ref{app:narration-pipeline}. The contribution of each design choice across all stages is isolated in \S\ref{sec:design-space}.

\begin{table*}[t]
\centering
\resizebox{0.87\textwidth}{!}{%
\small
\setlength{\tabcolsep}{4pt}
\begin{tabular}{@{}ll|rrrrr|rrrrr@{}}
\toprule
& & \multicolumn{5}{c|}{\textbf{DataTales}}
  & \multicolumn{5}{c}{\textbf{Earnings2Insights}} \\
\cmidrule(lr){3-7} \cmidrule(lr){8-12}
\textbf{Model} & \textbf{Method}
  & \%ins & \%ana & depth & \%fact & \%win
  & \%ins & \%ana & depth & \%fact & \%win \\
\midrule
\multirow{6}{*}{Qwen3-8B}
 & Direct  &  3.8 & 36.1 & 1.92 & 88.7 & 25.3
           &  6.6 & 38.8 & 2.04 & 94.9 & 88.6 \\
 & CoT     &  7.2 & 48.8 & 2.16 & \textbf{92.8} & 26.5
           &  4.2 & 38.3 & 2.15 & 94.5 & 58.8 \\
 & RAG     &  3.9 & 36.5 & 1.92 & 89.5 & 29.9
           &  4.4 & 34.0 & 1.95 & \textbf{96.1} & 87.8 \\
 & BoT     &  9.2 & 48.2 & 2.18 & 88.9 & 40.9
           &  5.2 & 31.3 & 1.55 & 91.8 & 86.3 \\
 & AWM     &  6.5 & 40.8 & 1.98 & 92.5 & 29.5
           &  5.0 & 39.9 & 1.99 & 93.1 & 80.2 \\
 & \method & \textbf{22.5} & \textbf{71.2} & \textbf{2.59}
           & 92.1 & \textbf{60.0}
           & \textbf{18.0} & \textbf{57.4} & \textbf{2.52}
           & 89.8 & \textbf{96.6} \\
\midrule
\multirow{6}{*}{Qwen3.5-9B}
 & Direct  &  5.2 & 31.1 & 1.79 & 93.6 & 54.9
           &  3.7 & 27.8 & 1.82 & 98.5 & 95.0 \\
 & CoT     &  5.7 & 35.5 & 1.84 & 95.2 & 51.6
           &  6.4 & 34.0 & 1.93 & 97.6 & 88.2 \\
 & RAG     &  8.5 & 44.7 & 2.04 & 94.2 & 70.2
           &  4.5 & 25.4 & 1.49 & 98.4 & 98.1 \\
 & BoT     & 14.0 & 50.4 & 2.19 & 95.8 & \textbf{88.2}
           &  3.2 & 19.1 & 1.66 & 98.8 & 74.6 \\
 & AWM     & 11.4 & 48.5 & 2.19 & \textbf{96.7} & 70.0
           &  6.4 & 37.7 & 2.05 & \textbf{98.9} & 97.3 \\
 & \method & \textbf{26.1} & \textbf{67.1} & \textbf{2.50}
           & 89.8 & 81.9
           & \textbf{23.8} & \textbf{63.1} & \textbf{2.61}
           & 97.2 & \textbf{99.6} \\
\midrule
\multirow{6}{*}{\shortstack[l]{DeepSeek\\-V4-Flash}}
 & Direct  &  5.5 & 36.7 & 1.79 & 94.0 & 75.6
           &  5.7 & 34.6 & 1.92 & \textbf{98.9} & 97.7 \\
 & CoT     &  6.6 & 43.7 & 1.98 & 95.0 & 83.1
           &  5.8 & 36.7 & 2.09 & 98.2 & 91.2 \\
 & RAG     &  7.9 & 46.9 & 2.10 & 92.4 & 85.4
           &  7.5 & 39.5 & 2.12 & 98.5 & 98.1 \\
 & BoT     & 12.7 & 55.6 & 2.27 & 90.8 & 88.3
           &  7.0 & 38.5 & 2.07 & 98.6 & 95.4 \\
 & AWM     & 10.5 & 52.4 & 2.25 & \textbf{96.1} & 96.3
           &  7.7 & 42.7 & 2.07 & 97.0 & 98.5 \\
 & \method & \textbf{21.1} & \textbf{67.6} & \textbf{2.48}
           & 90.9 & \textbf{99.3}
           & \textbf{18.1} & \textbf{55.5} & \textbf{2.36}
           & 97.7 & \textbf{100.0} \\
\midrule
\multirow{4}{*}{GPT-5.1}
 & Direct  &  9.8 & 54.6 & 2.17 & 94.8 & 71.1
           & 15.7 & 45.6 & 2.22 & 99.3 & \textbf{100.0} \\
 & CoT     &  4.0 & 28.8 & 1.57 & 94.8 & 43.3
           & 14.2 & 42.2 & 2.30 & \textbf{99.3} & \textbf{100.0} \\
 & RAG     & 10.5 & 60.1 & 2.34 & \textbf{95.9} & 90.1
           &  7.3 & 38.4 & 2.04 & 99.0 & 99.2 \\
 & \method & \textbf{22.3} & \textbf{64.9} & \textbf{2.43}
           & 95.5 & \textbf{99.6}
           & \textbf{16.9} & \textbf{53.5} & \textbf{2.34}
           & 98.5 & 99.2 \\
\bottomrule
\end{tabular}%
}
\caption{\textbf{Main results} across two benchmarks and four backbones. Columns: insight rate (\%ins), analysis rate (\%ana), reasoning depth (depth), factual precision (\%fact), and win rate (\%win). Insight rate is the headline metric; analysis rate is the control. Best per model in \textbf{bold}.}
\label{tab:main-results}
\vspace{-0.1cm}
\end{table*}

\section{Experiments}
\label{sec:setup}

\paragraph{Benchmarks.}
We evaluate on the two analytical report generation benchmarks. \textbf{DataTales}~\cite{yang2024datatales} pairs structured market data with expert daily reports (460 instances).  \textbf{Earnings2Insights}~\cite{takayanagi2025earnings2insights} provides  earnings call transcripts  for analytical report generation (132 instances, extended from the original 64).

\paragraph{Models.}
We test four backbones spanning scale and capability: \textbf{Qwen3-8B} and \textbf{Qwen3.5-9B} (small open-source), \textbf{DeepSeek-V4-Flash} (reasoning model), and \textbf{GPT-5.1} (proprietary), each in identical configurations across all conditions.

\paragraph{Baselines.}
We compare against two families. The prompting family consists  \textbf{Direct}, \textbf{CoT}, and \textbf{RAG}. The structural-level pattern family consists of Buffer of Thoughts (\textbf{BoT})~\cite{yang2024bot} and Agent Workflow Memory (\textbf{AWM})~\cite{wang2024awm}, which are close to
\method{} in approach: BoT retrieves thought templates and AWM
retrieves task-level workflows, both from a library at inference
time. Both are adapted to distill from the same 550-report corpus that \method consumes (adaptation details in Appendix~\ref{app:bot-awm-adaptation}).

\paragraph{Metrics.} 
We evaluate along five dimensions measuring the quality of analytical-level report generation. \textbf{Insight rate}, the fraction of claims that surface insights beyond what careful reading of the source data would yield (e.g., connecting a revenue decline to a peer's steeper decline to infer market share gain), is the headline metric. \textbf{Analysis rate}, the fraction of claims containing insights or standard expert analysis (e.g., decomposing that same revenue decline into volume and price components), measures the overall expert-level analytical content in the report. \textbf{Reasoning depth}, the mean analytical hops per claim, measures whether the gain comes from deeper inference chains. \textbf{Factual precision}, the fraction of numerical values correct against the source, ensuring analytical depth does not trade off correctness. \textbf{Win rate}, the rate at which the generated report is preferred over or tied with the expert reference, judged by two LLM personas (analyst, investor) under order-swapped trials. Judgments use Gemini-3-Flash; protocol details
and reliability checks are in Appendix~\ref{app:metric-reliability}, where the LLM judge's claim labels agree with human annotators at a level comparable to inter-annotator agreement, and human and judge insight rates correlate at the report level.

\begin{table*}[t]
\centering
\small
\setlength{\tabcolsep}{4.5pt}
\definecolor{posgreen}{RGB}{0, 128, 80}
\definecolor{negred}{RGB}{190, 70, 50}
\newcommand{\gain}[1]{\textcolor{posgreen}{\scriptsize(+#1)}}
\newcommand{\loss}[1]{\textcolor{negred}{\scriptsize(#1)}}
\begin{tabular}{@{}llccccc@{}}
\toprule
& \textbf{Configuration} & \%insight & \%analysis & depth & \%factual & \%win \\
\midrule
\textbf{Axis} & Full pipeline (default) & 23.8 & 63.1 & 2.61 & 97.2 & 99.6 \\
\midrule
\multirow{2}{*}{\textit{Representation}}
 & Single-field            & 20.3 \gain{3.5} & 58.1 \gain{5.0} & 2.50 \gain{0.11} & 98.2 \loss{-1.0} & 99.2 \gain{0.4} \\
 & Two-field               & 23.9 \loss{-0.1} & 62.1 \gain{1.0} & 2.57 \gain{0.04} & 96.6 \gain{0.6} & 97.6 \gain{2.0} \\
\midrule
\textit{Retrieval target}
 & Reference text          & 19.3 \gain{4.5} & 56.8 \gain{6.3} & 2.48 \gain{0.13} & 96.4 \gain{0.8} & 100.0 \loss{-0.4} \\
\midrule
\textit{Integration}
 & Skip per-Analysis gen.  & 19.2 \gain{4.6} & 57.9 \gain{5.2} & 2.52 \gain{0.09} & 98.1 \loss{-0.9} & 99.6 {\scriptsize\textcolor{gray}{(0.0)}} \\
\midrule
\multirow{3}{*}{\textit{Composition}}
 & Transcript only         & 13.0 \gain{10.8} & 49.1 \gain{14.0} & 2.39 \gain{0.22} & 98.6 \loss{-1.4} & 97.7 \gain{1.9} \\
 & Signals only            & 11.1 \gain{12.7} & 45.4 \gain{17.7} & 2.21 \gain{0.40} & 98.8 \loss{-1.6} & 96.6 \gain{3.0} \\
 & Findings only           & 25.1 \loss{-1.3} & 65.9 \loss{-2.8} & 2.70 \loss{-0.09} & 94.1 \gain{3.1} & 96.5 \gain{3.1} \\
\bottomrule
\end{tabular}
\caption{Ablation study on Earnings2Insights with Qwen3.5-9B. Each row changes one axis from the full pipeline. \textcolor{posgreen}{Green} deltas indicate where the full pipeline is better; \textcolor{negred}{red} deltas indicate where the variant is better.}
\label{tab:design-space}
\end{table*}

\section{Results and Discussions}


\paragraph{Main Results.}
\label{sec:main-results}

Table~\ref{tab:main-results} reports results across all conditions. \method lifts insight rate by 1.7--3.7$\times$ over all baselines. \method{} consistently outperforms all baselines: 

On DataTales, insight rate (first column) rises from a best-baseline range of 9.2--14.0\% to 21.1--26.1\%; on Earnings2Insights, from 6.4--7.7\% to 18.0--23.8\% for the three non-GPT backbones. Baselines cluster in a narrow band regardless of prompting strategy: on DataTales with Qwen3-8B, insight rate rises by 2.4--5.9$\times$, analysis rate by 1.3--2.1$\times$, and reasoning depth by 0.4--0.7 hops, while factual precision remains comparable (89--97\% across conditions) and win rate reaches 60--100\%.
Baselines cluster in a narrow band regardless of method: across all backbones, prompting methods fall within 4--16\% insight rate of each other, and the structural-level methods (BoT, AWM) do not escape this band despite distilling from the same 550-report corpus, indicating a ceiling on insight for methods operating at the structural level.

To verify that insight gains stem from the library mechanism rather than prompting effects, we analyze 30 instances with the largest insight-rate gap between \method{} and direct prompting. First, 80.4\% of \method{}'s novel claims trace to a retrieved \emph{Analysis} (44\% strongly, 37\% partially), confirming the library as the primary driver of novel content. Second, for 58\% of these claims, the baseline either omits the same signal entirely (25\%) or mentions it only generically (34\%). The remaining 42\% cite the same data but do not draw the analytical inference \method{} makes, illustrating the selection challenge: the data is available but the right analytical move is not applied. Table~\ref{tab:mechanism-example} shows a representative traced claim alongside the baseline's treatment of the same input. Third, \method{} covers 4.6 distinct signal types per instance vs.\ 0.83 for the baseline, a 5.5$\times$ gap. \method{} produces novel insights across a wider range of data signals rather than concentrating on the most prominent one.

\begin{table}[t]
\centering
\small
\setlength{\tabcolsep}{4pt}
\renewcommand{\arraystretch}{1.05}
\begin{tabular}{@{}p{1.15cm}p{6.3cm}@{}}
\toprule
\multicolumn{2}{@{}l@{}}{\textbf{Retrieved \emph{Analysis}}} \\
\addlinespace[2pt]
\emph{signal} & A company reports a very large amount of cash generated
after necessary investments, and this amount is growing rapidly, while
another well-known company with a much larger overall valuation generates
only moderately more of this same type of cash flow. \\
\addlinespace[3pt]
\emph{move} & Compare the company's cash generation level and growth to
that of a significantly larger peer by relating each company's cash flow
to its overall valuation, and use this comparison to assess how
efficiently the company converts its value into cash relative to the
peer. \\
\midrule
\multicolumn{2}{@{}l@{}}{\textbf{Report generated for} \texttt{PARA\_q1\_2022}} \\
\addlinespace[2pt]
Ours & ``Operating cash flow grew 34\% YoY to \$347M, but it represents
only 31.5\% of the peer's \$1.1B, suggesting lower cash flow
efficiency.'' \\
\addlinespace[3pt]
Direct & Does not cite operating cash flow figures or any peer
comparison. \\
\bottomrule
\end{tabular}
\caption{A novel claim traced to the \emph{Analysis} that produced it (Earnings2Insights, \texttt{PARA\_q1\_2022}).}
\label{tab:mechanism-example}
    \vspace{-0.3cm}
\end{table}

The gain is smallest for GPT-5.1 on Earnings2Insights, where Direct and CoT already reach 14--16\% insight rate natively, leaving less headroom for the library. On DataTales, where no analysis is stated in the raw price and volume input, GPT-5.1 still gains 2.1$\times$ (22.3\% vs.\ 10.5\%), matching the lift on weaker backbones.



\paragraph{Human Evaluation.}
We perform two human evaluations to check that LLM-based metrics reflect genuine quality. To validate insight quality, two human annotators independently ranked 30 report triplets (\method{}, best prompting baseline, BoT) by insight quality, blind to method identity. \method{} is preferred in 96.7--100\% of pairwise comparisons on Earnings2Insights and 63.3--86.7\% on DataTales, confirming that \method{} produces more insightful reports across both benchmarks. Annotators agree on which report ranks first in 66.7\% of sets, with most disagreements concerning the ordering of the two baselines, which independently confirm the structural-level ceiling observed in the automated metrics.

To validate factuality, a human audit of sampled claims checks the content beyond the numerical values covered by \%factual: \method{} contradicts the source in 4.7\% of claims on the stronger backbone, below the pooled baseline, with the higher rate on Qwen3-8B tracking the reasoning capability that valid inference requires. Protocols and full results for both evaluations are in Appendix~\ref{app:human-eval}.

\subsection{Ablation Study} 
\label{sec:design-space}

Table~\ref{tab:design-space} isolates the contribution of each design choice by changing one axis at a time from the full pipeline (Qwen3.5-9B, Earnings2Insights). 

\paragraph{Representation.}

Reducing to a single field (raw reference text only) drops insight rate by 3.5 points. Two analyses reveal why. First, at a cosine threshold of 0.85, the abstracted \texttt{data\_signal} index retrieves 20$\times$ more library entries than \texttt{reference\_text}, whose concrete prose produces a sparse embedding space with few retrievable neighbors. Second, 90\% of entries retrieved via \texttt{data\_signal} come from a different sector than the input, compared to 44\% for \texttt{reference\_text}, which clusters within the source domain. Abstraction makes the library globally reusable rather than locally bound. The two-field variant (fusing \texttt{data\_signal} and \texttt{analytical\_move} into one field) recovers most of the insight rate with a small drop on win rate (2 points), suggesting that separating the retrieval key from the generation instruction may further improves output quality.

\paragraph{Retrieval target.}
Retrieving on \texttt{reference\_text} instead of \texttt{data\_signal} drops insight rate by 4.5 points. Because \texttt{reference\_text}s are concrete prose, similarity is driven by surface wording rather than by whether the stored pattern fits the input. For example, the input ``E\&I delivered a 6\% volume increase'' retrieves an entry whose reference span cites ``a growth rate of 6\% annually''. The two share wording but not pattern: the analytical move retrieved converts forward expectations into an implied growth rate, while the input reports an actual change, so the move is not executable. Across reports, only 36.4\% of the moves retrieved on \texttt{reference\_text} are executed, against 52.8\% for \texttt{data\_signal}, which selects the more applicable move on 80\% of reports.

\paragraph{Integration strategy.}
Skipping Stage~3 (per-\emph{Analysis} generation) and passing retrieved \emph{Analyses} directly to composition drops insight rate by 4.6 points. The \texttt{analysis\_move} is an imperative that must be executed against the input's signals, not merely supplied as context; without this execution step, the composer defaults to the generic analytical moves that characterize the baseline ceiling.

\paragraph{Composition input.}
Removing findings (the Stage~3 outputs produced by executing
each retrieved analytical move) from Stage~4 halves insight rate to 11--13\%, while the findings-only condition preserves it at 25.1\%. This confirms that executed analytical moves, not raw input or the data signal extracted, derive insight content. The full pipeline adds signals and transcript for factual grounding: precision recovers from 94.1\% to 97.2\% and win rate from 96.5\% to 99.6\%, a deliberate trade-off for a modest 1.3-point insight cost.

\subsection{Cross-Domain Transferability}
\label{sec:cross-domain}


\begin{figure}[t]
    \centering
    \includegraphics[width=\columnwidth]{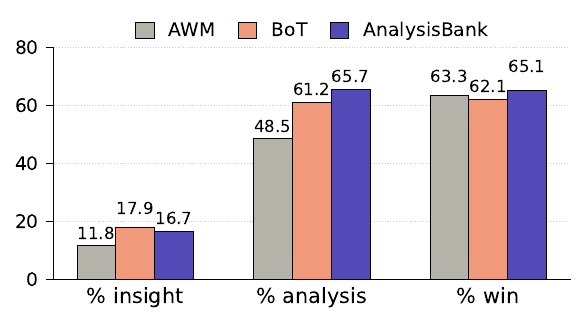}
    \vspace{-0.7cm}
    \caption{Cross-domain results on SciGen (Qwen3-8B). All three
    methods distill patterns from the same training corpus. Factual
    precision is comparable across methods (${\sim}$94\%).}
    \label{fig:cross-domain}
    \vspace{-0.3cm}
\end{figure}

The extraction pipeline and \emph{Analysis} representation are designed to be domain-agnostic: the move typology, abstraction procedure, and quality filter make no assumptions specific to finance. To test this, we apply \method to \textbf{SciGen}~\cite{Moosavi2021scigen}, a scientific paper generation benchmark, extracting \emph{Analyses} from its training set and evaluating with Qwen3-8B against BoT and AWM on the same corpus (Figure~\ref{fig:cross-domain}).

The pattern partially replicates. \method achieves the highest analysis rate (65.7\% vs.\ 61.2\% for BoT and 48.5\% for AWM) and win rate (65.1\%), confirming that the three-field decomposition transfers without architectural changes. On insight rate, BoT slightly edges \method (17.9\% vs.\ 16.7\%), a reversal from the financial benchmarks where \method leads by 1.7--3.7$\times$. We attribute this to the nature of scientific writing: the dominant analytical moves (methodology comparison, result interpretation, limitation identification) are fewer and more predictable than in financial analysis, reducing the long-tail advantage that drives \method's gains on DataTales and Earnings2Insights. BoT's full-report templates are better matched to this more uniform move distribution, while \method's per-\emph{Analysis} decomposition adds coverage in the tail that SciGen's narrower move vocabulary does not reward as strongly. The key result is nonetheless positive: the extraction pipeline produces a functional library from a non-financial corpus, and the resulting reports achieve the highest overall quality without domain-specific adaptation.

\section{Conclusion}


Analytical report generation requires more than deciding what a report should cover: it requires determining what analyses the data warrants. Existing methods largely leave this reasoning to model defaults, leading to generic analytical behavior despite differences in report structure. \method{} addresses this gap by retrieving and applying distilled expert \emph{Analyses} at inference time, substantially increasing the proportion of novel, data-grounded insights across benchmarks and model backbones. Its transfer to scientific writing further suggests that operating at the analytical level may generalize beyond finance to other forms of expert, data-grounded reasoning.

Two directions extend this work. First, the current pipeline applies each \emph{Analysis} independently in a single pass, but expert analysis is both iterative and compositional: one finding may trigger further Analyses, and some insights require chaining multiple moves into a compound finding. Iterative retrieval-generation~\citep{shao2023iterative,trivedi2023interleaving} and analysis composition~\citep{press2023measuring} offer natural mechanisms but remain unexplored for analytical generation. Second, expert reports pair analysis with visualizations, yet the current pipeline produces text only. Extending the \emph{Analysis} representation with visualization specifications~\citep{yang2025multimodaldeepresearcher} would enable multi-modal reports.

\section*{Limitations}
The quality of the \emph{Analysis} library is bounded by the expert reports it is distilled from. If the source reports are formulaic or lack deep analytical reasoning, the resulting \emph{Analyses} will reflect those limitations. While our quality filter mitigates some of this risk, the pipeline cannot produce analytical patterns richer than what the source corpus contains. Our experiments are conducted exclusively on English corpora. The extraction pipeline relies on LLM-based abstraction and generalization, and its effectiveness on other languages has not been tested. \method{} is designed for reports requiring data-specific analytical reasoning. For domains where a standard, fixed set of analyses suffices, the analytical-level approach offers limited advantage over structural-level methods, as the long-tail coverage that drives \method{}'s gains would not be needed.

\section*{Ethical Considerations} 
The expert reports used for \emph{Analysis} extraction are publicly available financial documents. We release only the distilled \emph{Analyses} and extraction code, not the source reports themselves, to respect the intellectual property of the original authors and publishers. 

Our automated evaluation relies on LLM-based judgment. Human evaluation involved two voluntary annotators from the authors' professional network who ranked system-generated reports without compensation; no personal data was collected. 

Generated reports may contain analytical errors or unsupported inferences. We strongly recommend human review before any investment decision is based on system-generated content, and advocate for human-AI collaboration where the system augments rather than replaces expert judgment. Any deployment of this technology should clearly disclose that reports are AI-generated to ensure transparency for readers and stakeholders.

\bibliography{custom}

\input{latex/appendix}

\end{document}

%% file: latex/appendix.tex
\appendix

\section{Extraction Pipeline Configuration}
\label{app:extraction-config}

This appendix details the hyperparameters, prompts, and implementation choices for the four-pass extraction pipeline (§\ref{sec:extraction}). The pipeline takes a CSV of expert analyst reports with columns \texttt{sector, symbol, company, author, date, url, cleaned\_text} and writes a SQLite-backed \emph{Analysis} library.

\subsection{Hyperparameters}
\label{app:extraction-hyperparams}

Pass A tags each identified move with one of seven seed types (attribution, derivation, flagging, comparison, projection, gap detection, and stress test) to steer extraction, 
together with three accompanying fields (\texttt{what\_analyst\_did}, \texttt{what\_triggered\_it}, \texttt{result\_text}).

\begin{table}[h]
\resizebox{\columnwidth}{!}{%
\centering
\small
\setlength{\tabcolsep}{6pt}
\begin{tabular}{@{}lll@{}}
\toprule
\textbf{Pass} & \textbf{Parameter} & \textbf{Value} \\
\midrule
Pass A (Induce)         & temperature                 & 0.3 \\
                        & move types                  & 7  \\
\midrule
Pass B (Generalize)     & temperature                 & 0.3 \\
                        & retry budget                & 1   \\
\midrule
Pass C (Deduplicate)    & temperature                 & 0.2 \\
                        & cluster threshold (cosine)  & 0.88 \\
                        & max cluster size            & 30  \\
                        & cross-batch dedup threshold & 0.85 \\
                        & embedding batch             & 100 \\
                        & clustering algorithm        & greedy single-linkage \\
\midrule
Pass D (Quality-filter) & temperature                 & 0.1 \\
                        & criteria                    & 3 (transferable, \\
                        &                             & actionable, grounded) \\
                        & retry budget                & 1   \\
\bottomrule
\end{tabular}%
}
\caption{Extraction pipeline hyperparameters.}
\label{tab:extraction-hyperparams}
\end{table}

\subsection{Pass A: Induce}
\label{app:extraction-pass1}

\begin{tcolorbox}[colback=gray!4,colframe=gray!50,boxrule=0.4pt,arc=2pt,
                  fontupper=\footnotesize\ttfamily,
                  title={\normalfont\textit{Pass A system prompt}}]
You are examining an expert financial analysis report to identify
analytical moves --- moments where the analyst performed a specific
reasoning operation rather than simply describing or summarizing.

An analytical move has three properties:

1. It required the analyst to do something active
   (compare, derive, decompose, flag, project, contrast)

2. It produced an insight not directly stated in the source material

3. It could in principle be applied to a different company or situation

For each analytical move you find, output:

- "move\_type": one word describing the operation
  [attribution | derivation | flagging | comparison | projection |
   gap\_detection | stress\_test]

- "what\_analyst\_did": one sentence describing the specific operation
  performed

- "what\_triggered\_it": what data condition or pattern in the source
  material prompted this move

- "result\_text": the exact span in the report that is the output of
  this move

Do not output moves that are simple description or summary.
Output only moves that are non-generic and could transfer to a
different earnings report.
Output only valid JSON, no commentary.
\end{tcolorbox}

Pass A issues one LLM call per report. The system prompt frames the task around three properties of an analytical move (active reasoning, non-trivial output, transferability) and requests a JSON list of moves. Malformed items missing any of the four required keys are silently dropped. The output is a per-report list of candidate moves; reports yielding no moves are skipped.

\subsection{Pass B: Generalize}
\label{app:extraction-pass2}

Pass B issues one LLM call per candidate. The system prompt requires a JSON object with exactly the three fields used by the rest of the pipeline (\texttt{data\_signal}, \texttt{analytical\_move}, \texttt{reference\_text}) and includes an explicit transferability test (``would this description match the same situation in a hospital, a retailer, and a defense contractor?''). The pass is callable with a \texttt{retry\_instruction} that is appended to the system prompt; this slot is used by Pass D to request targeted re-generation when a \emph{Analysis} fails the transferability or actionability check.

\begin{tcolorbox}[colback=gray!4,colframe=gray!50,boxrule=0.4pt,arc=2pt,
                  fontupper=\footnotesize\ttfamily,
                  title={\normalfont\textit{Pass B system prompt}}]
You are converting a specific analytical move from one expert report
into a reusable \emph{Analysis} applicable to any earnings report.

Output a JSON object with three fields:

- "data\_signal": the data pattern that should trigger this
  \emph{Analysis}. Written at the causal pattern level. No company names,
  specific numbers, or industry terminology. A reader should recognize
  this pattern in a report about any company in any sector.

- "analytical\_move": what to do when this pattern is found. Written
  as a clear imperative. Specific enough that two analysts following
  it independently would produce similar outputs.

- "reference\_text": copied verbatim from result\_text in the input.

Test for data\_signal quality: would this description match the
same situation in a hospital, a retailer, and a defense contractor?
If yes, the abstraction is correct. If it contains domain-specific
terms, abstract further.

Output only valid JSON, no commentary.
\end{tcolorbox}

Source metadata is attached deterministically after Pass B completes: \texttt{sector}, \texttt{symbol}, \texttt{company}, \texttt{date}, \texttt{url}, and \texttt{author} are copied from the CSV row, and the \texttt{reference\_text} is wrapped with a trailing source label ``\texttt{— <company> <date>}'' before being stored as the first element of the \texttt{reference\_texts} list.

\paragraph{Retry instructions used by Pass D.}
\begin{tcolorbox}[colback=amber!8,colframe=amber!60!black,boxrule=0.4pt,arc=2pt,
                  fontupper=\footnotesize\ttfamily,
                  title={\normalfont\textit{transferability retry}}]
The data\_signal is too domain-specific. Abstract it further so
it would match the same causal pattern in a hospital, a retailer,
and a defense contractor. Remove all industry terminology and
specific numbers.
\end{tcolorbox}

\begin{tcolorbox}[colback=amber!8,colframe=amber!60!black,boxrule=0.4pt,arc=2pt,
                  fontupper=\footnotesize\ttfamily,
                  title={\normalfont\textit{actionability retry}}]
The analytical\_move is too vague. Replace any phrase like
"analyze X" with specific operations: decompose, quantify, attribute,
flag, or project. Name what to decompose into, what to quantify,
and what to compare against.
\end{tcolorbox}

\subsection{Pass C: Deduplicate}
\label{app:extraction-pass3}

Pass C has two stages: a deterministic clustering step on embeddings and an LLM merge step per cluster. Candidates are embedded by their \texttt{data\_signal} using \texttt{text-embedding-ada-002} in batches of 100. Clustering is greedy single-linkage on the unit-normalized embeddings: each candidate is assigned to the first existing cluster whose centroid has cosine similarity $\geq 0.88$ with the candidate; otherwise a new cluster is opened. Centroids are running means re-normalized after each addition. Clusters larger than 30 candidates are split into chunks of 30 before being sent to the LLM merge step, to keep prompts within context.

The LLM merge step receives the cluster as a JSON payload of \texttt{(data\_signal, analytical\_move, reference\_texts)} triples and is asked to return a merged \emph{Analysis} per cluster. Reference texts are taken verbatim from the originating candidates rather than from the LLM output, since the LLM was observed to occasionally paraphrase what the prompt required to be verbatim spans. Provenance for a merged \emph{Analysis} is reconstructed by matching the merged \texttt{data\_signal} back to the originating candidates and aggregating per-field via a scalar-or-list rule (single distinct value $\to$ scalar; multiple $\to$ list).

For incremental updates against an existing library, Pass C runs a cross-batch deduplication step before clustering: each new candidate is compared by cosine similarity against existing \emph{Analysis} embeddings, and candidates above the cross-batch threshold of 0.85 are paired with their nearest existing \emph{Analysis} and sent through Pass C as a targeted merge that preserves the existing \emph{Analysis}'s ID (so downstream stores update rather than insert).

\begin{tcolorbox}[colback=gray!4,colframe=gray!50,boxrule=0.4pt,arc=2pt,
                  fontupper=\footnotesize\ttfamily,
                  title={\normalfont\textit{Pass C merge prompt}}]
You are given a list of candidate \emph{Analyses}. Identify groups of
\emph{Analyses} that represent the same analytical operation at the pattern
level, even if their reference\_texts come from different reports or
industries.

For each group:

- Keep the data\_signal that is most general while still being
  specific enough to trigger reliably

- Keep the analytical\_move that is most complete and specific

- Combine all reference\_texts into an array --- multiple examples
  strengthen the validation anchor. Each entry should end with
  a source label e.g. "--- Boeing Q3 2022"

Return \emph{Analyses} that do not match any other as singletons.
Output the merged \emph{Analysis} library as a JSON array.
Each element must have exactly these fields: data\_signal,
analytical\_move, reference\_texts.
Output only valid JSON, no commentary.
\end{tcolorbox}

\subsection{Pass D: Quality-filter}
\label{app:extraction-pass4}

\begin{tcolorbox}[colback=gray!4,colframe=gray!50,boxrule=0.4pt,arc=2pt,
                  fontupper=\footnotesize\ttfamily,
                  title={\normalfont\textit{Pass D system prompt}}]
Evaluate this \emph{Analysis} against three criteria.
For each, output "pass" or "fail" and one sentence of reasoning.

TRANSFERABLE: would data\_signal trigger on the same pattern
in a different industry? Fail if it contains domain-specific terms.

ACTIONABLE: if two analysts followed analytical\_move independently,
would they produce similar outputs? Fail if the action is vague
(e.g. "analyze X" rather than "decompose X into Y and Z").

GROUNDED: does reference\_text show a real instance of the pattern
and its output, not just a description of what the pattern is?

Output JSON with fields: transferable, actionable, grounded,
each containing "result" (pass|fail) and "reason" (string).
A \emph{Analysis} must pass all three to enter the library.
\end{tcolorbox}

Pass D issues one LLM call per \emph{Analysis} asking for a pass/fail verdict on each of three criteria. The criteria are evaluated independently, but the routing of failures is asymmetric:

\begin{itemize}
\item \textbf{Grounded fail} $\to$ discard. A \emph{Analysis} whose reference span does not show the pattern executed cannot be repaired by re-generation, so it is removed from the library.
\item \textbf{Transferable fail} $\to$ retry Pass B with the transferability retry instruction (above). The originating moves are re-sent through Pass B with the appended instruction, re-merged via Pass C, and re-evaluated. If the retry passes all three criteria, the resulting \emph{Analysis} enters the library with the original ID preserved; otherwise it is discarded.
\item \textbf{Actionable fail} $\to$ retry Pass B with the actionability retry instruction. Same loop as above.
\end{itemize}

Parse failures on the Pass D response are treated conservatively as pass-all rather than discard-all, on the principle that an inability to judge a \emph{Analysis} is not evidence against it; this affects $<1\%$ of \emph{Analyses} in practice.

\subsection{Concurrency and Caching}
\label{app:extraction-concurrency}

The pipeline parallelizes at two levels: rows of the input CSV are processed in parallel (one thread per row), and within each row Pass B calls are issued in parallel across the moves Pass A identified. All LLM-call results, embeddings, and intermediate candidates are written to a SQLite store keyed by \emph{Analysis} ID and embedding model identifier, so re-runs of the pipeline skip already-extracted \emph{Analyses} and already-embedded items. Embeddings for clustering and for cross-batch dedup share the same cache.

\section{Bank Details}
\label{app:bank-details}


\begin{figure*}[p]
    \centering
    \includegraphics[width=\textwidth]{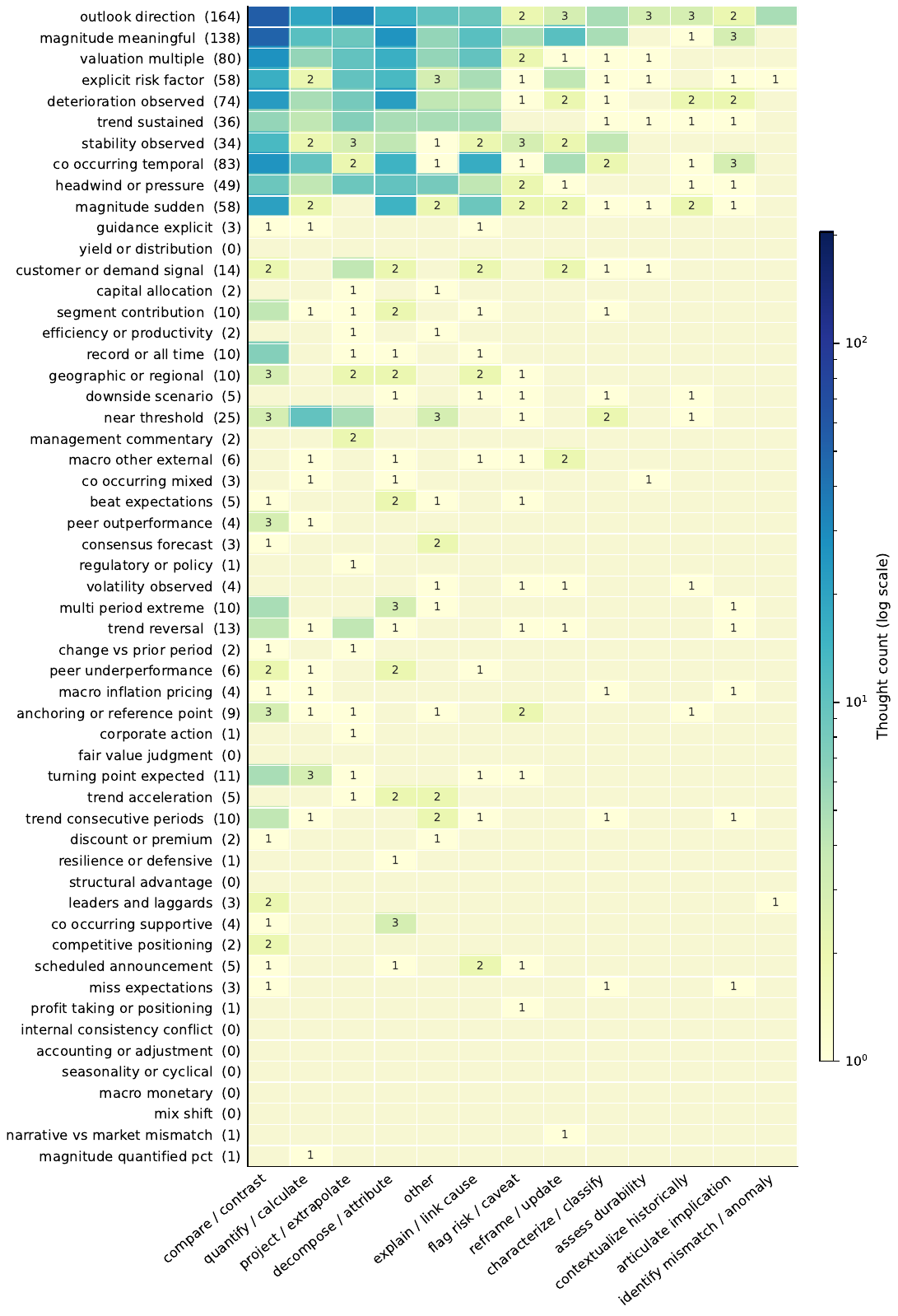}
    \vspace{-0.8cm}
    \caption{Distribution of \emph{Analyses} across signal types
    and analytical moves in the DataTales library
    ($n{=}1{,}422$). Rows are signal types and columns are
    analytical move types, both ordered by total frequency across
    the two libraries so that this figure and
    Figure~\ref{fig:bank-heatmap-earnings} share axes and color
    scale. Cell color encodes \emph{Analysis} count on a log
    scale; sparse cells (counts 1--3) are annotated with their
    value, and absent pairs are shaded grey. Numbers in
    parentheses are per-library signal totals.}
    \label{fig:bank-heatmap-datatales}
\end{figure*}

\begin{figure*}[p]
    \centering
    \includegraphics[width=\textwidth]{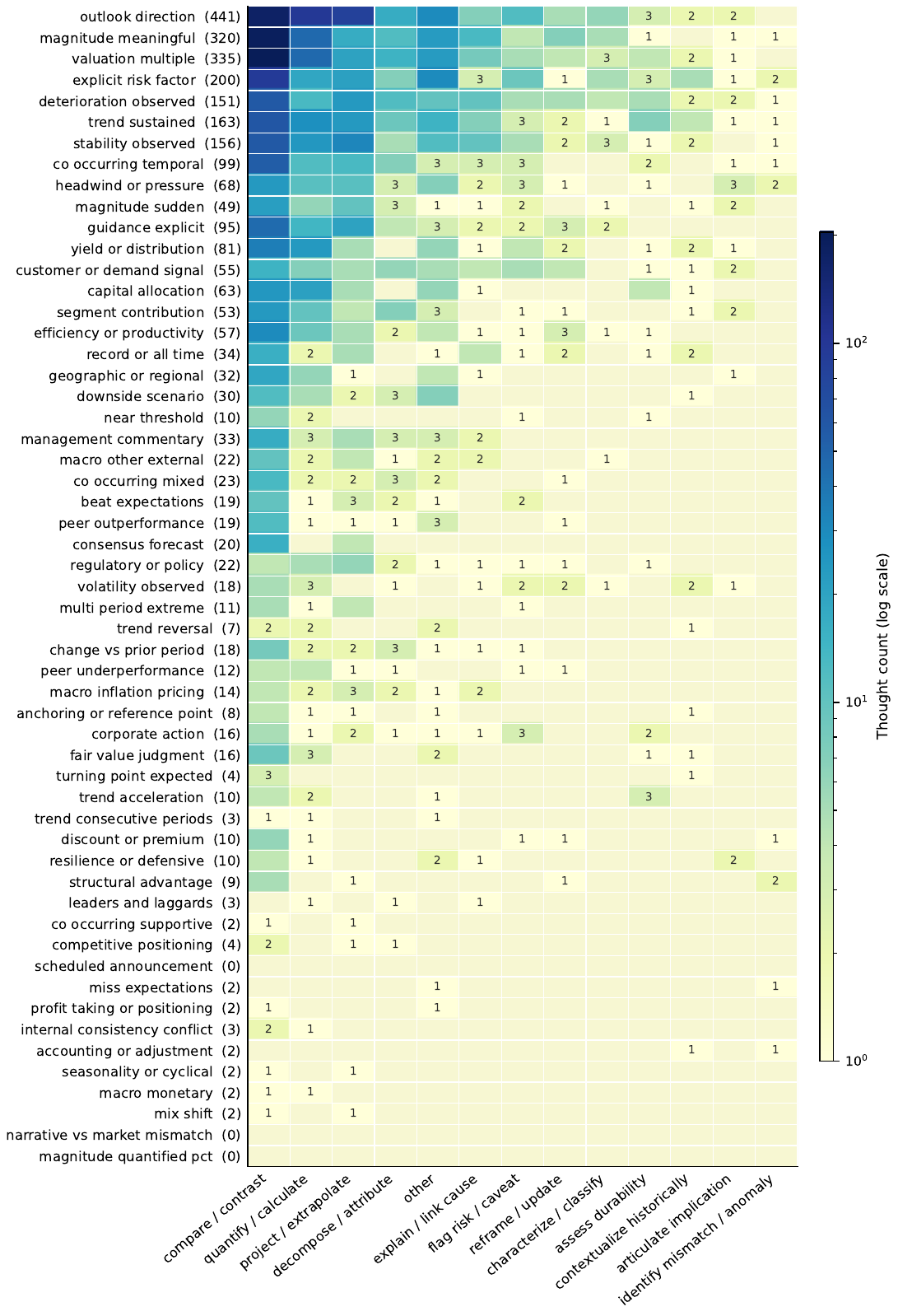}
    \caption{Distribution of \emph{Analyses} across signal types
    and analytical moves in the Earnings library
    ($n{=}3{,}889$). Axes, ordering, and color scale are shared
    with Figure~\ref{fig:bank-heatmap-datatales}. Both libraries
    concentrate on the same high-frequency signal-move
    combinations while exhibiting a long tail of rare but distinct
    analytical patterns.}
    \label{fig:bank-heatmap-earnings}
\end{figure*}

We characterize each library along two axes, using one keyword taxonomy per axis, since signals are noun-phrase data conditions and moves are verb-phrase operations. Each taxonomy is a hand-built, ordered list of categories defined by keyword patterns over the \texttt{data\_signal} and \texttt{analytical\_move} fields, refined against the libraries until coverage stabilized; we assign each \emph{Analysis} to the first matching category (single-label) and collect non-matches as \emph{other}. 
Figures~\ref{fig:bank-heatmap-datatales} and~\ref{fig:bank-heatmap-earnings} show the full signal-by-move distribution for both corpora.

\section{Narration Pipeline Configuration}
\label{app:narration-pipeline}

This appendix details the hyperparameters (Table~\ref{tab:narration-hyperparams}), prompts, and implementation choices for the four-stage narration pipeline (§\ref{sec:narration-pipeline}). The pipeline takes a source input (an earnings call transcript with sector tag, or a structured OHLCV DataFrame with a report date and market tag) and produces an analyst-style report.


\begin{table}[h]
\resizebox{\columnwidth}{!}{%
\centering
\small
\setlength{\tabcolsep}{6pt}
\begin{tabular}{@{}lll@{}}
\toprule
\textbf{Stage} & \textbf{Parameter} & \textbf{Value} \\
\midrule
\multicolumn{3}{l}{\textit{Stage 1 (signal extraction)}} \\
\quad Earnings (LLM)    & temperature                    & 0.2 \\
                        & mode                           & hierarchical (default) \\
                        & signal types                   & 17 (14 presence + 3 absence) \\
                        & per-type cap                   & 6--12 (see Table~\ref{tab:hier-caps}) \\
\quad Earnings (validation) & temperature                & 0.1 \\
                        & error classes                  & 6 \\
\quad DataTales (pandas) & lookback windows              & 1d, 5d, 20d \\
                        & momentum threshold             & 0.5\% \\
                        & vol spike / collapse           & 1.5$\times$ / 0.5$\times$ hist \\
                        & volume spike threshold         & 2.0$\times$ 20d avg \\
                        & MA threshold                   & 0.5\% from 20d MA \\
\midrule
\multicolumn{3}{l}{\textit{Stage 2 (retrieval)}} \\
\quad Top-$k$           & per input                      & 5 \\
\quad Retrieval mode    & default                        & per-type \\
\quad Retrieval backend & default                        & \texttt{text-embedding-ada-002} \\
\quad Sector boost      & multiplicative                 & 1.2$\times$ on match \\
\quad Similarity threshold &                             & 0.0 (no floor) \\
\quad Triggers per \emph{Analysis}  &                            & up to 2 \\
\midrule
\multicolumn{3}{l}{\textit{Stage 3 (per-\emph{Analysis} analysis)}} \\
\quad Temperature       &                                & 0.4 \\
\quad Mode              & default                        & pattern \\
\quad Parallelism       &                                & one thread per \emph{Analysis} \\
\midrule
\multicolumn{3}{l}{\textit{Stage 4 (composition)}} \\
\quad Temperature       &                                & 0.5 \\
\quad Mode              & default (Earnings)             & signals (transcript-free) \\
                        & default (DataTales)            & findings + signals + summary \\
\midrule
\multicolumn{3}{l}{\textit{Validate-and-retry}} \\
\quad Temperature       &                                & 0.1 \\
\quad Quality verdicts  &                                & high / partial / missing \\
\quad Retry trigger     &                                & quality = missing \\
\quad Retry budget      &                                & 1 \\
\bottomrule
\end{tabular}%
}
\caption{Narration pipeline hyperparameters.}
\label{tab:narration-hyperparams}
\end{table}

\subsection{Stage 1: Signal extraction}
\label{app:narration-stage1}

The pipeline supports two signal-extraction backends, selected by input modality. Both produce the same output type: a list of \texttt{(type, fields, span)} tuples that drive Stage 2 retrieval.

\begin{tcolorbox}[colback=gray!4,colframe=gray!50,boxrule=0.4pt,arc=2pt,
                  fontupper=\footnotesize\ttfamily,
                  title={\normalfont\textit{Stage 1 prompt (one call per signal type)}}]
You are a financial signal extractor. Extract signals of
exactly ONE type from an earnings transcript.

SIGNAL TYPE: \{signal\_type\}

REQUIRED FIELDS:
\{fields\}

EXTRACTION RULE:
\{rule\}

OUTPUT FORMAT:
Return a JSON array. Each element must have EXACTLY these
fields:
\{
  "type": "\{signal\_type\}",
  "fields": \{ <the required fields above, omit fields with
             no evidence> \},
  "span": "<shortest exact verbatim quote from the
           transcript, 10-40 words, that supports this
           signal>"
\}

Rules:
- Extract EVERY distinct instance of this signal type ---
  do not stop at one.
- span must be copied verbatim from the transcript.
- For ABSENT types: span is the quote containing the
  claim/indicator that reveals the absence.
- Do not infer fields not supported by the transcript text.
- Return [] if no signals of this type are present.
- Output only a valid JSON array, no commentary.
\end{tcolorbox}

\paragraph{Earnings transcripts (LLM-based).}
For transcripts, an LLM extracts 17 signal types: 14 \emph{presence} signals (margin delta, incremental margin, guidance revision, volume trend, market share, pricing realization, FX impact, forward signal, cash flow, capital allocation, earnings quality, balance sheet, segment mix, management tone) and 3 \emph{absence} signals (absent mechanism, absent estimate, absent conversion). The hierarchical mode (default) issues one focused LLM call per signal type and runs them in parallel; each call is gated by a keyword-evidence check (e.g.\ the \texttt{MARGIN\_DELTA} call fires only if "margin", "basis point", "bps", or "profitability" appears in the transcript), with absence types having empty keyword lists so they always fire.

\paragraph{Per-type caps and numeric prioritization.}
Hierarchical extraction is permissive (the per-type prompt asks for \emph{every} distinct instance), so each type has a post-extraction cap to prevent any one type from saturating the slate. Within a capped type, signals with non-empty numeric fields (e.g.\ \texttt{magnitude\_pct} for \texttt{VOLUME\_TREND}, \texttt{delta\_bps} for \texttt{FX\_IMPACT}) are kept first; non-numeric signals fill remaining slots.

\begin{table}[h]
\resizebox{\columnwidth}{!}{%
\centering
\small
\setlength{\tabcolsep}{6pt}
\begin{tabular}{@{}llll@{}}
\toprule
Type & Cap & Type & Cap \\
\midrule
GUIDANCE\_REVISION   & 12 & MGMT\_TONE          & 8 \\
MARGIN\_DELTA        & 12 & PRICING\_REALIZATION & 8 \\
VOLUME\_TREND        & 10 & SEGMENT\_MIX        & 8 \\
FX\_IMPACT           & 10 & BALANCE\_SHEET      & 8 \\
MARKET\_SHARE        & 10 & EARNINGS\_QUALITY   & 8 \\
FORWARD\_SIGNAL      & 10 & ABSENT\_ESTIMATE    & 8 \\
CAPITAL\_ALLOCATION  & 10 & ABSENT\_CONVERSION  & 6 \\
ABSENT\_MECHANISM    & 10 & CASH\_FLOW          & 6 \\
                     &    & INCREMENTAL\_MARGIN & 6 \\
\bottomrule
\end{tabular}%
}
\caption{Per-type signal caps in hierarchical extraction.}
\label{tab:hier-caps}
\end{table}

\paragraph{Fact validation.}
After extraction, every signal is passed through a fact-validation LLM call that checks six error classes against the verbatim supporting span: \emph{geo\_scope} (region-specific figure attributed to the global segment), \emph{period\_scope} (full-year forecast tagged as a quarterly actual), \emph{metric\_identity} (segment-level figure labelled company-wide; guidance figure labelled actual), \emph{composite\_split} (combined dividend + buyback amount assigned to one action), \emph{organic\_vs\_reported} (reported growth labelled organic), and \emph{adjusted\_vs\_gaap} (non-GAAP figure labelled GAAP). Errors with a confident correction are applied in place; errors without are flagged. The validation prompt is conservative (only flags errors with clear span evidence), and parse failures are treated as pass-through.

\paragraph{DataTales (deterministic).}
For structured market data, signals are computed directly from OHLCV series in pandas with no LLM involvement. The extractor produces eight signal types: \texttt{PRICE\_MOMENTUM} (1d/5d/20d returns above a 0.5\% threshold), \texttt{VOLATILITY\_SIGNAL} (annualized 20d realized vol versus 60d rolling baseline, classified as spike / elevated / normal / suppressed), \texttt{VOLUME\_ANOMALY} (today's volume versus 20d average), \texttt{TREND\_SIGNAL} (close versus 20d MA), \texttt{TERM\_STRUCTURE} (front-vs-second-month spread for futures markets), \texttt{CONTRACT\_ROLL} (5d-average volume crossover), \texttt{YIELD\_CURVE} (2s10s, 2s30s, 3m10s spreads for treasury markets), and \texttt{CROSS\_ASSET} (VIX regime and 10Y vs equities for equity markets, USD direction for currencies, Brent-WTI spread for energy). For the equity market only, per-product signals are restricted to a set of primary instruments (S\&P 500, Nasdaq Composite, Nasdaq 100, Dow Jones, Russell 2000, VIX, US 10-Year Bond Yield) to prevent the slate from being saturated by individual constituents.

\paragraph{Signal validation (span check).}
For LLM-extracted signals, the supporting span is verified against the transcript via whitespace-normalized substring match; signals whose span cannot be located are flagged \texttt{span\_hallucinated} and dropped. \texttt{MARGIN\_DELTA} cause-types are additionally checked against the supporting span; unsupported causes are stripped from the signal rather than flagging it whole.

\subsection{Stage 2: Retrieval}
\label{app:narration-stage2}

Stage 2 retrieves a slate of $k=5$ \emph{Analyses} from \method given the signal list. Three retrieval modes and three backends are supported, with the defaults indicated below.

\paragraph{Embedding target.}
The default retrieves by cosine similarity between signal descriptions and the \texttt{data\_signal} field of each \emph{Analysis}. An ablation retrieves against \texttt{reference\_texts} (concatenated with trailing source attributions stripped); this variant is reported in §\ref{sec:design-space}.

\paragraph{Sector boost.}
Each \emph{Analysis}'s similarity score is multiplied by 1.2 if the \emph{Analysis}'s source sector matches the input's; sector membership is checked as exact match or as set-membership when a \emph{Analysis} has multiple source sectors. The boost favors in-sector \emph{Analyses} without hard-filtering out cross-sector ones, since some analytical moves transfer across sectors.

\paragraph{Retrieval mode.}
The default is \emph{per-type} retrieval: each fired signal type contributes one representative signal (selected by largest numeric magnitude using a per-type priority list — \texttt{delta\_bps} for margin signals, \texttt{magnitude\_pct} for volume, both \texttt{old\_guidance} and \texttt{new\_guidance} for guidance, etc.), and the best \emph{Analysis} for that representative is added to the slate. Types are processed in order of representative-signal significance, and remaining slate slots are filled by global cosine top-$k$ to avoid leaving the slate short when fewer than $k$ signal types fire. The alternative \emph{cosine} mode applies global top-$k$ directly without per-type structure.

\paragraph{Retrieval backend.}
The default backend embeds signals and \emph{Analyses} with  OpenAI's \texttt{text-embedding-ada-002}. Two alternatives are available for ablation: BM25 over tokenized signal and \emph{Analysis} text, and a local sentence-transformer (\texttt{all-MiniLM-L6-v2}). Backend choice is independent of retrieval mode (e.g.\ BM25 + per-type is valid). All backends apply the same 1.2$\times$ sector boost and the same optional similarity-threshold cutoff (default 0, i.e.\ no filtering on raw similarity).

\paragraph{Trigger mapping.}
For each \emph{Analysis} in the final slate, the top-2 signals by cosine similarity to the \emph{Analysis} embedding are tagged as the \emph{Analysis}'s \emph{triggering signals}; their verbatim spans are passed to Stage 3 alongside the \emph{Analysis}.

\paragraph{Embedding cache.}
\emph{Analysis} embeddings are computed once and stored in a SQLite cache keyed by \emph{Analysis} ID and embedding model. The default \texttt{data\_signal} embeddings and the reference-text variant each live in their own cache table, so switching ablation variants does not require re-embedding.

\subsection{Stage 3: Per-\emph{Analysis} analysis}
\label{app:narration-stage3}

For each \emph{Analysis} in the Stage 2 slate, an independent LLM call applies the \emph{Analysis}'s \texttt{analytical\_move} to its triggering signals and supporting spans (transcript-side) or to the market data context (DataTales). The prompt forbids section headers and framing language, because Stage 4 imposes structure. Calls are issued in parallel across \emph{Analyses} via a thread pool.

The user prompt assembles three blocks: \texttt{PATTERN}, the \texttt{analytical\_move} to execute; \texttt{SIGNALS}, the triggering signals serialized as JSON (\texttt{type} and structured \texttt{fields}); and \texttt{EXCERPTS}, their verbatim supporting spans.

\begin{tcolorbox}[colback=gray!4,colframe=gray!50,boxrule=0.4pt,arc=2pt,
                  fontupper=\footnotesize\ttfamily,
                  title={\normalfont\textit{Stage 3 prompt}}]
You are a financial analyst performing a specific analysis.
Apply the analytical pattern to the provided signals and
transcript excerpts.
Be precise and quantitative where the data supports it.

Do not write section headers or introductory framing.
Your output will be combined with other analyses into a
broader report.

---

PATTERN:
\{analytical\_move\}

SIGNALS:
[\{"type": ..., "fields": \{...\}\}, ...]

EXCERPTS:
"\{verbatim supporting span\}"
...
\end{tcolorbox}

\paragraph{Three reference modes (ablation).}
The default mode (\texttt{pattern}) sends only the pattern, signals, and spans, omitting the \emph{Analysis}'s \texttt{reference\_text}. The \texttt{pattern\_with\_refs} mode additionally includes the reference text(s) as a depth and tone anchor, inserted as a \texttt{REFERENCE EXAMPLES} block between \texttt{PATTERN} and \texttt{SIGNALS}. The \texttt{refs\_only} mode passes only the reference text(s) and signals, without the pattern — testing whether reference examples alone are sufficient. The ablation isolates which component of the three-field representation carries the analytical depth gain (§\ref{sec:design-space}).

\paragraph{DataTales adaptation.}
For DataTales, Stage 3 uses a market-specific system prompt that forbids corporate-earnings language ("management", "guidance", "EPS") and asks for prices, spreads, and time-window references instead. Triggering signals are still the top-2 by cosine, but the supporting context is the deterministic market data summary (§\ref{app:narration-data-summary}) rather than transcript spans.

\subsection{Stage 4: Composition}
\label{app:narration-stage4}

A single LLM call composes the per-\emph{Analysis} findings into a final report. The prompt enforces three structural constraints: an executive summary (3--4 sentences naming takeaway, bull case, and key risk), theme sections with analyst-chosen titles drawn from the evidence, and an investment-recommendations block with horizon-differentiated calls (Next Day / Week / Month) each tagged Long / Short / Neutral with a one- to two-sentence rationale. The prompt directs the writer to treat findings as the analytical backbone and to use signals only for factual gap-filling.

\begin{tcolorbox}[colback=gray!4,colframe=gray!50,boxrule=0.4pt,arc=2pt,
                  fontupper=\footnotesize\ttfamily,
                  title={\normalfont\textit{Stage 4 prompt}}]
You are writing an earnings analysis report.
The following analytical findings have been derived from the
earnings call. Compose them into a coherent report organised
around investment-relevant themes that cut across multiple
metrics (e.g., demand environment, margin dynamics, capital
allocation, guidance credibility, regional mix, risk
factors). Let the data shape the themes --- use as many
sections as the story requires.

Do not repeat analysis --- synthesize and connect where
findings relate to each other.
Use the signals to fill any factual gaps the findings do not
cover.

REPORT STRUCTURE:

**Executive Summary** (3-4 sentences)
State the single most important takeaway, the dominant bull
case, and the key risk. This should read as the opening of a
research note --- sharp and investment-actionable.

**[Theme sections --- analyst's choice of titles]**
Each section covers a cross-cutting investment theme. For
each:
- Lead with the analytical insight that frames the theme.
- Weave in specific numbers, percentages, and period
  comparisons from the relevant findings.
- Where evidence from multiple findings reinforces or
  conflicts with each other, integrate it.
- Close each section with the forward implication for the
  investment case.

**Investment Recommendations**
After synthesising the full picture, provide differentiated
calls across horizons. Each horizon reflects a distinct lens
--- do not recycle the same rationale:
- **Next Day:** Headline beat/miss reaction and any guidance
  surprise.
- **Next Week:** Sector/macro regime and near-term catalysts
  or overhangs.
- **Next Month:** Fundamental execution quality --- margins,
  cash flow, order book durability.
For each horizon state: **Long / Short / Neutral** and a 1-2
sentence rationale naming the specific data points that drove
the call.
\end{tcolorbox}

\paragraph{Input variants.}
Four prompt variants are used depending on what context Stage 4 receives:
\begin{itemize}
\item \emph{Default (findings + signals)}: the production configuration. Signals fill factual gaps the findings do not cover.
\item \emph{With transcript}: additionally supplies the raw transcript or market data summary, with a primary rule that the findings still drive structure and the transcript only fills coverage gaps. Used when the deployment can tolerate the additional input length.
\item \emph{Transcript-only / signals-only / findings-only}: ablations that strip the input progressively, used in the component analysis in §\ref{sec:design-space}.
\end{itemize}
For DataTales, the same four variants exist with adapted system prompts that swap earnings vocabulary for market-data vocabulary and that ask for trading-relevant horizon calls (momentum continuation, technical structure, term-structure outlook) rather than equity-analyst horizons.

\paragraph{DataTales context augmentation.}
On DataTales, Stage 4 prompts may optionally include a \emph{per-entity raw OHLCV block} (the last 20 rows for each product on or before the report date). This is included automatically in the default mode and in the signals-only ablation when a DataFrame is provided to the pipeline. It is not used in the transcript-equivalent earnings flow.

\subsection{Validate-and-retry}
\label{app:narration-validate}

After composition, a per-\emph{Analysis} validator checks whether each \emph{Analysis}'s analytical pattern was applied in the report. The validator issues one LLM call per \emph{Analysis} in the slate (in parallel) and emits a JSON verdict with \texttt{applied} (boolean), \texttt{quality} ("high", "partial", or "missing"), and a one-sentence note. Verdicts of \texttt{missing} trigger a targeted re-run: Stage 3 is re-executed for only the missing \emph{Analyses}, the updated findings are merged into the existing findings list, and Stage 4 is re-composed. The loop is bounded by a retry budget (default 1) and terminates as soon as no missing verdicts remain.

The validator uses the same backbone LLM as the rest of the pipeline. Parse failures default to \texttt{missing=False, quality="missing"}, which is conservative in the sense that it does not trigger spurious retries but does flag the \emph{Analysis} as not validated.

\subsection{Market data summary (DataTales only)}
\label{app:narration-data-summary}

For DataTales inputs, a deterministic summarizer converts the OHLCV DataFrame into a human-readable context block used by Stages 3 and 4. For each product on or before the report date, the summary emits: last close, 1d/5d/20d returns, position relative to the 20d moving average, annualized 20d realized volatility, and volume ratio against the 20d average. For futures markets, a term-structure line is appended (front vs.\ second-month spread with backwardation/contango label). For the equity market, primary instruments get a full paragraph each; secondary products are collapsed into a one-line "sector / other products" summary capped at ten entries.

\subsection{Concurrency and caching}
\label{app:narration-concurrency}

Stage 1 hierarchical extraction, Stage 3 per-\emph{Analysis} analysis, and validate-stage checks all use thread pools sized to the number of items processed (one thread per signal type, per \emph{Analysis}, or per validation call). LLM responses are cached at the level of \texttt{(prompt, system\_prompt, temperature)} tuples; cached calls are served without an API hit, which lets ablations re-use intermediate results across configurations that share earlier stages. \emph{Analysis} embeddings are persisted to SQLite as described in §\ref{app:narration-stage2}.

The full default configuration on Earnings2Insights requires roughly $1 + 17 + 1 + 5 + 1 + 5 = 30$ LLM calls per report ($1$ if hierarchical mode condenses to a single call, $17$ for the per-type fan-out, $1$ for fact validation, $5$ for Stage 3, $1$ for Stage 4, $5$ for the validator) before any retries; on DataTales it is $5 + 1 + 5 = 11$ since Stage 1 is deterministic. Retries add up to $5 + 1$ calls per cycle, bounded by the retry budget.

\section{Structure-Level Baseline Adaptation}
\label{app:bot-awm-adaptation}

Buffer of Thoughts (BoT)~\cite{yang2024bot} and Agent Workflow Memory (AWM)~\cite{wang2024awm} were originally evaluated on math reasoning and web-agent trajectories respectively. We adapt both to analytical report generation while preserving their native abstraction granularity (one full-report template or workflow per task), so that the comparison with \method (\S\ref{sec:setup}) isolates decomposition granularity as the variable under test. Both methods are induced from the same 550-report corpus that \method consumes, with the same backbone model for induction and inference.

\subsection{Agent Workflow Memory}
\label{app:awm-adaptation}

AWM induces reusable workflows from training trajectories and injects the relevant workflow into the inference-time prompt.

\paragraph{Trajectory representation.}
Each expert report in the induction corpus is decomposed into a sequence of \texttt{(observation, action)} pairs by splitting on section headings (markdown headings, bold headers, or colon-terminated section labels). Each pair represents one reasoning step the analyst took: the heading is the \texttt{observation} (the context the analyst was about to write into), and the section body is the \texttt{action} (what was written). Reports without recognisable section structure are split into fixed 400-character chunks. This mirrors AWM's original trajectory format from web-agent execution traces.

\paragraph{Offline induction.}
Workflows are induced per GICS sector (eleven sectors for Earnings2Insights; per-market for DataTales). For each sector, up to ten training reports are sampled, decomposed into trajectories, and passed to an LLM with an induction prompt that asks for a 5--10 step numbered workflow capturing the analytical pattern shared across the reports. The induction prompt enforces abstraction (``do not reproduce specific figures from these reports'') and reusability (``apply to any future report for this ticker/sector''). One workflow per sector is stored as a plain-text file in a sector-keyed directory.

\paragraph{Inference-time retrieval.}
For a new input with ticker $t$ in sector $s$, the workflow is resolved by a four-step fallback: (i)~exact ticker match, (ii)~exact sector match, (iii)~LLM-selected closest peer from same-sector candidates, (iv)~LLM-selected cross-sector fallback. Step (iii) uses a structured JSON prompt over a ranked menu of candidate workflows; step (iv) is reached only when sector metadata is unavailable. In our benchmark configuration, step (ii) resolves every test instance, so the LLM-selection fallback is rarely exercised.

\paragraph{Inference-time generation.}
The selected workflow is injected at the top of the generation prompt as an ``Analyst Workflow (induced from past reports)'' block, followed by the task description, the source data, and a request to ``follow the workflow above as a guide for structuring your analysis [and] adapt each step to the current data.'' One LLM call produces the final report. Crucially, the workflow is the only retrieved artifact: AWM does not extract typed signals, does not retrieve multiple patterns, and does not run per-pattern execution before composition.

\subsection{Buffer of Thoughts}
\label{app:bot-adaptation}

BoT maintains a meta-buffer of structured ``thought templates'' and applies a four-step inference pipeline (problem distillation $\to$ buffer retrieval $\to$ template instantiation $\to$ reasoner instantiation) per task. We use the canonical 5-section template format from the original paper: \textit{Key Information}, \textit{Domain Constraints}, \textit{Abstract Task Description}, \textit{Python Logic}, and \textit{Answer Format}.

\paragraph{Offline induction.}
Templates are induced per training report rather than per sector. For each report in the induction set (sampled at ten per sector to match AWM's induction budget), the report is passed to an LLM with a thought-distillation prompt that asks for a 5-section template covering the analytical pattern the report instantiates, with all company-specific figures abstracted away. The resulting template is compared against the closest existing template in the buffer via an LLM-judged novelty check; templates judged novel are added, near-duplicates are skipped. The novelty filter prevents the buffer from accumulating paraphrases of the same template while still allowing genuinely distinct analytical patterns to coexist. Templates are stored in a persistent ChromaDB collection with sentence-transformer embeddings (all-MiniLM-L6-v2) of the template text.

\paragraph{Inference-time retrieval.}
For a new input, an LLM first performs \emph{problem distillation}: it produces a 4-section distillation of the input (Key Information, Domain Constraints, Abstract Task Description, Answer Format) that serves as the retrieval query. Cosine similarity against the meta-buffer returns the single closest template. The retrieved template is then \emph{instantiated} by a second LLM call, which adapts each abstract section to the current company, period, and data, producing a problem-specific briefing.

\paragraph{Inference-time generation.}
The instantiated template is injected at the top of the generation prompt as an ``Analyst Briefing (Buffer of Thought)'' block, followed by the task description and the source data. One LLM call produces the final report. As with AWM, the template is the only retrieved artifact: a single top-down structure covers the whole report, and no per-move retrieval or per-pattern execution occurs.

\paragraph{Online buffer update.}
The original BoT formulation supports an optional online update that distils each newly produced report back into a template and adds it to the buffer if a novelty check passes. We disable this in the evaluated configuration to keep the buffer fixed and comparable to AWM's offline-only induction.

\subsection{Why this adaptation is fair}

We preserve each method's native abstraction granularity so that decomposition granularity is the variable under test. Adapting BoT or AWM to retrieve at the move level would convert them into variants of \method rather than independent baselines. We do adapt both to consume the same induction corpus as \method (\S\ref{sec:extraction}), so any performance gap reflects how the corpus is decomposed, not what it contains. The resulting contrast is explicit: at induction, each report yields one workflow or template versus the multiple analytical moves \method extracts; at inference, one structure is retrieved top-down versus the slate of move-level patterns \method retrieves and composes.

\section{Metric}
\label{app:metric-reliability}

This appendix specifies the judge models, the per-metric judgment protocols, and the design choices that bear on reliability.

\subsection{Judge models}

Two LLM judges are used in the evaluation pipeline:

\begin{itemize}
\item \textbf{Theme extractor (Claude Sonnet 4.6).} Reads each source input (transcript or market-data summary) once, independently of any generated report, and emits the list of \emph{material themes} — themes a competent analyst would be expected to address. The same theme list is then reused across every model variant evaluated on that input, so coverage scoring is anchored to a fixed reference rather than re-elicited per condition.
\item \textbf{Per-claim and per-report judge (Gemini 3 Flash).} Runs claim extraction (Stage 1), reasoning-depth scoring (Stage 2a), data-analytic-style scoring (Stage 2b), theme-coverage matching, factual scoring, and head-to-head preference scoring against expert references. Every judgment uses temperature 0.0 and structured JSON output, with up to three retries on JSON-parse failure.
\end{itemize}

Both judges are different model families from any of the four evaluation backbones (Qwen3-8B, Qwen3.5-9B, DeepSeek-V4-Flash, GPT-5.1) to reduce self-preference bias in line with recent LLM-as-judge findings.

\subsection{Per-metric protocols}

\paragraph{\%insight and \%analysis.}
The judge first extracts analytical claims from the report in 80-character-minimum chunks split on paragraph breaks, with boilerplate (executive summary, investment-recommendations sections) stripped before extraction. Each claim is then labelled in one of four categories:
\begin{itemize}
\item \textbf{factual}: depth-1 claims that restate a number or fact from the source.
\item \textbf{novel}: a data-specific insight that combines conditions, attributes causes, or projects consequences in a way not directly stated in the source.
\item \textbf{standard}: an analytical claim that a competent analyst would routinely make from this data.
\item \textbf{generic}: a claim that applies to any report of this type and does not depend on the specific data.
\end{itemize}
\%insight is $\text{novel} / (\text{novel} + \text{standard} + \text{generic} + \text{factual})$; \%analysis adds standard to the numerator.

\paragraph{depth.}
For each extracted claim the judge counts \emph{analytical hops} — distinct inference steps needed to derive the claim from the source — and emits the chain of hop types (e.g.\ \texttt{FACTUAL}, \texttt{COMPARE}, \texttt{ATTRIBUTE}, \texttt{PROJECT}). depth is the mean hop count per claim. Depth-1 claims are auto-assigned the \emph{factual} label and skipped from analytical type scoring, so the two metrics share a single claim-extraction pass and are not double-counted.

\paragraph{\%factual.}
Numerical values in the report are checked sentence-by-sentence against the source. For Earnings2Insights the reference is the transcript itself: for each report sentence containing a financial number, candidate reference sentences are selected from the transcript by content-token overlap (stopwords stripped), and the judge labels each numerical value as \texttt{CORRECT}, \texttt{INCORRECT}, or \texttt{DONT\_KNOW}. For DataTales the reference is a structured table of pre-computed values (OHLCV plus derived metrics: 1d/5d/20d returns, 20d moving average, annualized volatility, volume ratio, term-structure spread); every report sentence is checked against the full table without a token-overlap filter, since every entry is potentially relevant. Date/time tokens (years, quarters, ISO dates, day-of-month integers adjacent to month names) are filtered out before scoring so they do not contaminate the financial-value count. \%factual is $\text{correct} / (\text{correct} + \text{incorrect})$ over the full report.

\paragraph{\%win.}
Each model report is judged head-to-head against the expert reference (Seeking Alpha analyst reports for Earnings2Insights, the dataset's human-written market reports for DataTales) by two personas — \emph{analyst} and \emph{investor} — each with its own system prompt and rubric. To control for position bias, each persona judges the pair under both orderings (model first, expert first), yielding $2 \times 2 = 4$ judgments per report. The two ordering outcomes are aggregated per persona by majority: two agreeing outcomes set the persona's verdict; \emph{model+tie} resolves to model and \emph{expert+tie} to expert, treating ties as the weaker signal in either direction; \emph{model+expert} resolves to tie. \%win is reported per persona and as the mean across personas.

\subsection{Reliability checks}

To test the claim-level labels against human judgment, two human annotators applied the LLM judge's claim taxonomy (novel/standard/generic/factual) to 277 claims (10 instances, \method{} and the best baseline), blind to both system identity and the LLM judge's labels. Table~\ref{tab:claim-agreement} reports the results. Annotators mark \method{} claims as novel significantly more often than baseline claims (15.3\% vs.\ 1.7\%) and as analytical significantly more often (51.0\% vs.\ 25.4\%), and human and judge per-report rates correlate for both metrics (Pearson $r{=}0.64$ and $0.65$). On the full-taxonomy taxonomy, the judge agrees with each annotator ($\kappa{=}0.41$ and $0.47$) at a level comparable to the annotators' agreement with each other ($\kappa{=}0.50$). Agreement is lowest on the novel-vs.-rest cut ($\kappa{=}0.25$): annotating a single claim as novel is subjective, so we treat the report-level rates as the unit of evaluation.

\begin{table}[t]
\resizebox{\columnwidth}{!}{%
\centering
\small\begin{tabular}{@{}lcc@{}}
\toprule
& \textbf{\%novel} & \textbf{\%analysis} \\
\midrule
Human rate: \method{} (\%) & 15.3 & 51.0 \\
Human rate: baseline (\%) & 1.7 & 25.4 \\
$\Delta$ (pp) [95\% CI] & $+13.7$ {\footnotesize [8.7, 18.3]} & $+25.6$ {\footnotesize [17.8, 33.7]} \\
Report-level Pearson $r$ & 0.64 & 0.65 \\
Claim-level $\kappa$ (inter-annotator) & 0.25 & 0.49 \\
\midrule
Full-taxonomy $\kappa$: LLM judge--annotator & \multicolumn{2}{c}{0.41, 0.47} \\
Full-taxonomy $\kappa$: inter-annotator & \multicolumn{2}{c}{0.50} \\
\bottomrule
\end{tabular}%
}
\caption{Claim-level human study (277 claims, blind to system and LLM judge labels). Claim-level $\kappa$ is inter-annotator agreement on each metric's cut (novel vs.\ rest; novel$+$standard vs.\ rest); full-taxonomy $\kappa$ gives agreement on the full taxonomy.}
\label{tab:claim-agreement}
\end{table}

\section{Human Evaluation Details}
\label{app:human-eval}

\subsection{Insight quality}

\paragraph{Sampling.} 
For each benchmark, we rank all Qwen3-8B test instances by the insight-rate gap between \method{} and the best baseline, then stratify-sample 5 instances from each tercile (top, middle, bottom), yielding 15 sets per benchmark (30 total). Each set contains three reports: \method{}, the best prompting baseline (CoT for DataTales, Direct for Earnings2Insights), and BoT.

\paragraph{Protocol.} 
Two annotators judge each set. Reports are labelled A, B, C with method identity hidden and presentation order randomized per set. Annotators rank the three reports from 1 (most insightful) to 3 (least insightful), with no ties allowed. The instruction defines insight as ``substantive, non-obvious analytical content---e.g., specific causal explanations, comparisons, projections, or implications grounded in the input---rather than generic commentary, restated facts, or boilerplate language.''

\paragraph{Results by benchmark.}
Table~\ref{tab:human-eval} reports consensus pairwise outcomes (both annotators combined).

\begin{table}[t]
\resizebox{\columnwidth}{!}{%
\centering
\small
\begin{tabular}{@{}llccc@{}}
\toprule
\textbf{Benchmark} & \textbf{Comparison} & \textbf{Win} &
\textbf{Loss} & \textbf{Win\%} \\
\midrule
E2I & \method{} vs Direct & 29 & 1 & 96.7 \\
E2I & \method{} vs BoT & 30 & 0 & 100.0 \\
DataTales & \method{} vs CoT & 26 & 4 & 86.7 \\
DataTales & \method{} vs BoT & 19 & 11 & 63.3 \\
\bottomrule
\end{tabular}%
}
\caption{Human pairwise preferences (consensus across two
annotators, $n{=}30$ votes per comparison). No ties were
recorded.}
\label{tab:human-eval}
\end{table}

\paragraph{Inter-annotator agreement.} 

Table~\ref{tab:iaa-ci} reports inter-annotator agreement with bootstrap 95\% CIs ($N{=}30$). Annotators agree on the top-ranked report in 66.7\% of sets (chance: 33.3\%) and on the pairwise preferences in 90.0\% (vs.\ prompting) and 70.0\% (vs.\ BoT) of comparisons; agreement on the middle rank (36.7\%) is near chance. Cohen's $\kappa$ (0.37 vs.\ prompting; 0.13 vs.\ BoT) is lower than raw agreement because \method{} wins most comparisons and $\kappa$ is deflated under skewed label distributions~\cite{feinstein1990high}.

\begin{table}[t]
\resizebox{\columnwidth}{!}{%
\centering
\small
\begin{tabular}{@{}lcc@{}}
\toprule
\textbf{Statistic} & \textbf{Estimate} & \textbf{95\% CI} \\
\midrule
Rank-1 agreement & 66.7\% & $[48.8, 80.8]$ \\
Rank-2 agreement & 36.7\% & $[21.9, 54.5]$ \\
Rank-3 agreement & 50.0\% & $[33.2, 66.8]$ \\
Exact full-ranking match & 33.3\% & $[19.2, 51.2]$ \\
Mean Spearman $\rho$ & 0.483 & $[0.283, 0.667]$ \\
\midrule
Raw agreement (vs.\ prompting) & 90.0\% & $[74.4, 96.5]$ \\
Raw agreement (vs.\ BoT) & 70.0\% & $[52.1, 83.3]$ \\
\midrule
Win: \method{} vs.\ prompting & 91.7\% & $[83.3, 98.3]$ \\
Win: \method{} vs.\ BoT & 81.7\% & $[71.7, 91.7]$ \\
\bottomrule
\end{tabular}%
}
\caption{Inter-annotator agreement and pairwise preference statistics with bootstrap 95\% CIs ($N{=}30$). Chance level for the rank agreements is 33.3\%.}
\label{tab:iaa-ci}
\end{table}

\subsection{Factuality}
We complement \%factual with a human evaluation on a small sample covering both numerical and non-numerical claims: 564 claims, sampled as in the human--judge agreement study (Table~\ref{tab:claim-agreement}) across \method{} and the baselines on Qwen3-8B and DeepSeek-V4-Flash, are each labelled \emph{supported}, \emph{contradicted} by the source, or \emph{unverifiable} (blind to system identity; annotator fixed across backbones). We report the contradiction rate (Table~\ref{tab:claim-validity}); the baselines' novel claims are too few for a meaningful rate and are skipped. On the stronger backbone, \method{}'s contradiction rate is low (4.7\% over all claims; 9.4\% on novel claims) and below the pooled baseline (6.6\%). On Qwen3-8B it is higher (15.3\%; 28.6\% on novel claims) while the baseline is stable (5.5\%): valid inference tracks the reasoning capability of the backbone, most strongly for novel claims. Most novel claims are unverifiable against the source alone (61--69\%), as they state content the input does not.

\begin{table}[t]
\resizebox{\columnwidth}{!}{%
\centering
\small
\begin{tabular}{@{}lccc@{}}
\toprule
& \multicolumn{2}{c}{\method{}} & pooled baseline \\
\cmidrule(lr){2-3}
Backbone & all claims & novel claims & all claims \\
\midrule
Qwen3-8B & 15.3\% (23/150) & 28.6\% (8/28) & 5.5\% (7/127) \\
DeepSeek-V4-Flash & 4.7\% (7/150) & 9.4\% (3/32) & 6.6\% (9/137) \\
\bottomrule
\end{tabular}%
}
\caption{Human factual audit of generated claims: contradiction rate (contradicted claims / audited claims), annotator fixed across backbones.}
\label{tab:claim-validity}
\end{table}

%% file: custom.bib
@article{wu2023bloomberggpt,
  title={BloombergGPT: A Large Language Model for Finance},
  author={Shijie Wu and Ozan Irsoy and Steven Lu and Vadim Dabravolski and Mark Dredze and Sebastian Gehrmann and Prabhanjan Kambadur and David Stuart Rosenberg and Gideon Mann},
  journal={ArXiv},
  year={2023},
  volume={abs/2303.17564},
  url={https://api.semanticscholar.org/CorpusID:257833842}
}

@article{yang2023fingpt,
  title={FinGPT: Open-Source Financial Large Language Models},
  author={Hongyang Yang and Xiao-Yang Liu and Chris Wang},
  journal={ArXiv},
  year={2023},
  volume={abs/2306.06031},
  url={https://api.semanticscholar.org/CorpusID:259129734}
}

@article{islam2023financebench,
  title={FinanceBench: A New Benchmark for Financial Question Answering},
  author={Pranab Islam and Anand Kannappan and Douwe Kiela and Rebecca Qian and Nino Scherrer and Bertie Vidgen},
  journal={ArXiv},
  year={2023},
  volume={abs/2311.11944},
  url={https://api.semanticscholar.org/CorpusID:265294665}
}

@inproceedings{nitarach2025fincot,
    title = "{F}in{C}o{T}: Grounding Chain-of-Thought in Expert Financial Reasoning",
    author = "Nitarach, Natapong  and
      Sirichotedumrong, Warit  and
      Pitchayarthorn, Panop  and
      Taveekitworachai, Pittawat  and
      Manakul, Potsawee  and
      Pipatanakul, Kunat",
    editor = "Chen, Chung-Chi  and
      Winata, Genta Indra  and
      Rawls, Stephen  and
      Das, Anirban  and
      Chen, Hsin-Hsi  and
      Takamura, Hiroya",
    booktitle = "Proceedings of The 10th Workshop on Financial Technology and Natural Language Processing",
    month = nov,
    year = "2025",
    address = "Suzhou, China",
    publisher = "Association for Computational Linguistics",
    url = "https://aclanthology.org/2025.finnlp-2.8/",
    doi = "10.18653/v1/2025.finnlp-2.8",
    pages = "93--123"
}

@article{wang2024voyager,
  author       = {Guanzhi Wang and
                  Yuqi Xie and
                  Yunfan Jiang and
                  Ajay Mandlekar and
                  Chaowei Xiao and
                  Yuke Zhu and
                  Linxi Fan and
                  Anima Anandkumar},
  title        = {Voyager: An Open-Ended Embodied Agent with Large Language Models},
  journal      = {Trans. Mach. Learn. Res.},
  volume       = {2024},
  year         = {2024},
  url          = {https://openreview.net/forum?id=ehfRiF0R3a},
  bibsource    = {dblp computer science bibliography, https://dblp.org}
}

@inproceedings{finder2025choi,
  author       = {Chanyeol Choi and
                  Jihoon Kwon and
                  Jaeseon Ha and
                  Hojun Choi and
                  Chaewoon Kim and
                  Yongjae Lee and
                  Jy{-}yong Sohn and
                  Alejandro Lopez{-}Lira},
  title        = {FinDER: Financial Dataset for Question Answering and Evaluating Retrieval-Augmented
                  Generation},
  booktitle    = {{ICAIF}},
  url          = {https://doi.org/10.1145/3768292.3770361},
  pages        = {638--646},
  publisher    = {{ACM}},
  year         = {2025}
}

@inproceedings{wang2024trove,
  author       = {Zhiruo Wang and
                  Graham Neubig and
                  Daniel Fried},
  title        = {TroVE: Inducing Verifiable and Efficient Toolboxes for Solving Programmatic
                  Tasks},
  booktitle    = {{ICML}},
  series       = {Proceedings of Machine Learning Research},
  url          = {https://proceedings.mlr.press/v235/wang24az.html},
  pages        = {51177--51191},
  publisher    = {{PMLR} / OpenReview.net},
  year         = {2024}
}

@inproceedings{yang2024datatales,
  author       = {Yajing Yang and
                  Qian Liu and
                  Min{-}Yen Kan},
  title        = {DataTales: {A} Benchmark for Real-World Intelligent Data Narration},
  booktitle    = {{EMNLP}},
  url          = {https://aclanthology.org/2024.emnlp-main.601/"},
  pages        = {10764--10788},
  publisher    = {Association for Computational Linguistics},
  year         = {2024}
}

@inproceedings{takayanagi2025earnings2insights,
    title = "{E}arnings2{I}nsights: Analyst Report Generation for Investment Guidance",
    author = "Takayanagi, Takehiro  and
      Goldsack, Tomas  and
      Izumi, Kiyoshi  and
      Lin, Chenghua  and
      Takamura, Hiroya  and
      Chen, Chung-Chi",
    editor = "Chen, Chung-Chi  and
      Winata, Genta Indra  and
      Rawls, Stephen  and
      Das, Anirban  and
      Chen, Hsin-Hsi  and
      Takamura, Hiroya",
    booktitle = "FinNLP",
    month = nov,
    year = "2025",
    publisher = "Association for Computational Linguistics",
    url = "https://aclanthology.org/2025.finnlp-2.17/",
    pages = "246--251"
}

@inproceedings{wang2024awm,
  author       = {Zora Zhiruo Wang and
                  Jiayuan Mao and
                  Daniel Fried and
                  Graham Neubig},
  title        = {Agent Workflow Memory},
  booktitle    = {{ICML}},
  url          = {https://proceedings.mlr.press/v267/wang25bx.html},
  series       = {Proceedings of Machine Learning Research},
  publisher    = {{PMLR} / OpenReview.net},
  year         = {2025}
}

@inproceedings{yang2024bot,
  author       = {Ling Yang and
                  Zhaochen Yu and
                  Tianjun Zhang and
                  Shiyi Cao and
                  Minkai Xu and
                  Wentao Zhang and
                  Joseph E. Gonzalez and
                  Bin Cui},
  title        = {Buffer of Thoughts: Thought-Augmented Reasoning with Large Language
                  Models},
  url          = {http://papers.nips.cc/paper\_files/paper/2024/hash/cde328b7bf6358f5ebb91fe9c539745e-Abstract-Conference.html},
  booktitle    = {NeurIPS},
  year         = {2024}
}

@inproceedings{zhou2024selfdiscover,
  author       = {Pei Zhou and
                  Jay Pujara and
                  Xiang Ren and
                  Xinyun Chen and
                  Heng{-}Tze Cheng and
                  Quoc V. Le and
                  Ed H. Chi and
                  Denny Zhou and
                  Swaroop Mishra and
                  Huaixiu Steven Zheng},
  title        = {{SELF-DISCOVER:} Large Language Models Self-Compose Reasoning Structures},
  url          = {http://papers.nips.cc/paper\_files/paper/2024/hash/e41efb03e20ca3c231940a3c6917ef6f-Abstract-Conference.html},
  booktitle    = {NeurIPS},
  year         = {2024}
}

@article{jeong2025total,
  author       = {Soyeong Jeong and
                  Taehee Jung and
                  Sung Ju Hwang and
                  Joo{-}Kyung Kim and
                  Dongyeop Kang},
  title        = {When Thoughts Meet Facts: Reusable Reasoning for Long-Context LMs},
  journal      = {CoRR},
  volume       = {abs/2510.07499},
  year         = {2025},
  url          = {https://doi.org/10.48550/arXiv.2510.07499},
  doi          = {10.48550/ARXIV.2510.07499},
  eprinttype   = {arXiv},
  eprint       = {2510.07499},
  bibsource    = {dblp computer science bibliography, https://dblp.org}
}

@inproceedings{xie2023pixiu,
  author       = {Qianqian Xie and
                  Weiguang Han and
                  Xiao Zhang and
                  Yanzhao Lai and
                  Min Peng and
                  Alejandro Lopez{-}Lira and
                  Jimin Huang},
  title        = {{PIXIU:} {A} Comprehensive Benchmark, Instruction Dataset and Large
                  Language Model for Finance},
  url          = {http://papers.nips.cc/paper\_files/paper/2023/hash/6a386d703b50f1cf1f61ab02a15967bb-Abstract-Datasets\_and\_Benchmarks.html},
  booktitle    = {NeurIPS},
  year         = {2023}
}

@inproceedings{chen2021finqa,
  author       = {Zhiyu Chen and
                  Wenhu Chen and
                  Charese Smiley and
                  Sameena Shah and
                  Iana Borova and
                  Dylan Langdon and
                  Reema Moussa and
                  Matt Beane and
                  Ting{-}Hao Kenneth Huang and
                  Bryan R. Routledge and
                  William Yang Wang},
  title        = {FinQA: {A} Dataset of Numerical Reasoning over Financial Data},
  url          = {https://doi.org/10.18653/v1/2021.emnlp-main.300},
  booktitle    = {{EMNLP} {(1)}},
  pages        = {3697--3711},
  publisher    = {Association for Computational Linguistics},
  year         = {2021}
}

@inproceedings{zhu2021tatqa,
  author       = {Fengbin Zhu and
                  Wenqiang Lei and
                  Youcheng Huang and
                  Chao Wang and
                  Shuo Zhang and
                  Jiancheng Lv and
                  Fuli Feng and
                  Tat{-}Seng Chua},
  title        = {{TAT-QA:} {A} Question Answering Benchmark on a Hybrid of Tabular
                  and Textual Content in Finance},
  url          = {https://doi.org/10.18653/v1/2021.acl-long.254},
  booktitle    = {{ACL/IJCNLP} {(1)}},
  pages        = {3277--3287},
  publisher    = {Association for Computational Linguistics},
  year         = {2021}
}

@article{adfcot2025,
  author       = {Anmol Singhal Navya Singhal},
  title        = {Analogy-Driven Financial Chain-of-Thought (AD-FCoT): {A} Prompting
                  Approach for Financial Sentiment Analysis},
  journal      = {CoRR},
  volume       = {abs/2509.12611},
  year         = {2025},
  url          = {https://doi.org/10.48550/arXiv.2509.12611},
  doi          = {10.48550/ARXIV.2509.12611},
  eprinttype   = {arXiv},
  eprint       = {2509.12611},
  bibsource    = {dblp computer science bibliography, https://dblp.org}
}

@inproceedings{mukherjee2022ectsum,
  author       = {Rajdeep Mukherjee and
                  Abhinav Bohra and
                  Akash Banerjee and
                  Soumya Sharma and
                  Manjunath Hegde and
                  Afreen Shaikh and
                  Shivani Shrivastava and
                  Koustuv Dasgupta and
                  Niloy Ganguly and
                  Saptarshi Ghosh and
                  Pawan Goyal},
  title        = {ECTSum: {A} New Benchmark Dataset For Bullet Point Summarization of
                  Long Earnings Call Transcripts},
  url          = {https://doi.org/10.18653/v1/2022.emnlp-main.748},
  booktitle    = {{EMNLP}},
  pages        = {10893--10906},
  publisher    = {Association for Computational Linguistics},
  year         = {2022}
}

@article{augmentedllm2024,
  author       = {Van{-}Duc Le},
  title        = {Auto-Generating Earnings Report Analysis via a Financial-Augmented
                  {LLM}},
  journal      = {CoRR},
  volume       = {abs/2412.08179},
  year         = {2024},
  url          = {https://doi.org/10.48550/arXiv.2412.08179},
  doi          = {10.48550/ARXIV.2412.08179},
  eprinttype   = {arXiv},
  eprint       = {2412.08179},
  bibsource    = {dblp computer science bibliography, https://dblp.org}
}

@inproceedings{wiseman2017challenges,
  author       = {Sam Wiseman and
                  Stuart M. Shieber and
                  Alexander M. Rush},
  title        = {Challenges in Data-to-Document Generation},
  url          = {https://doi.org/10.18653/v1/d17-1239},
  booktitle    = {{EMNLP}},
  pages        = {2253--2263},
  publisher    = {Association for Computational Linguistics},
  year         = {2017}
}

@inproceedings{parikh2020totto,
  author       = {Ankur P. Parikh and
                  Xuezhi Wang and
                  Sebastian Gehrmann and
                  Manaal Faruqui and
                  Bhuwan Dhingra and
                  Diyi Yang and
                  Dipanjan Das},
  title        = {ToTTo: {A} Controlled Table-To-Text Generation Dataset},
  url          = {https://doi.org/10.18653/v1/2020.emnlp-main.89},
  booktitle    = {{EMNLP} {(1)}},
  pages        = {1173--1186},
  publisher    = {Association for Computational Linguistics},
  year         = {2020}
}

@inproceedings{yang2022re3,
  author       = {Kevin Yang and
                  Yuandong Tian and
                  Nanyun Peng and
                  Dan Klein},
  title        = {Re3: Generating Longer Stories With Recursive Reprompting and Revision},
  url          = {https://doi.org/10.18653/v1/2022.emnlp-main.296},
  booktitle    = {{EMNLP}},
  pages        = {4393--4479},
  publisher    = {Association for Computational Linguistics},
  year         = {2022}
}

@inproceedings{puduppully2019content,
  author       = {Ratish Puduppully and
                  Li Dong and
                  Mirella Lapata},
  title        = {Data-to-Text Generation with Content Selection and Planning},
  url          = {https://doi.org/10.1609/aaai.v33i01.33016908},
  booktitle    = {{AAAI}},
  pages        = {6908--6915},
  publisher    = {{AAAI} Press},
  year         = {2019}
}

@inproceedings{goldsack2025facts,
  author       = {Tomas Goldsack and
                  Yang Wang and
                  Chenghua Lin and
                  Chung{-}Chi Chen},
  title        = {From Facts to Insights: {A} Study on the Generation and Evaluation
                  of Analytical Reports for Deciphering Earnings Calls},
  url          = {https://aclanthology.org/2025.coling-main.705/},
  booktitle    = {{COLING}},
  pages        = {10576--10593},
  publisher    = {Association for Computational Linguistics},
  year         = {2025}
}

@article{koshkin2025margen,
  author       = {Roman Koshkin and
                  Pengyu Dai and
                  Nozomi Fujikawa and
                  Masahito Togami and
                  Marco Visentini Scarzanella},
  title        = {MaRGen: Multi-Agent {LLM} Approach for Self-Directed Market Research
                  and Analysis},
  journal      = {CoRR},
  volume       = {abs/2508.01370},
  year         = {2025},
  url          = {https://doi.org/10.48550/arXiv.2508.01370},
  doi          = {10.48550/ARXIV.2508.01370},
  eprinttype   = {arXiv},
  eprint       = {2508.01370},
  bibsource    = {dblp computer science bibliography, https://dblp.org}
}

@inproceedings{yang2025kahan,
  author       = {Yajing Yang and
                  Tony Deng and
                  Min{-}Yen Kan},
  title        = {{KAHAN:} Knowledge-Augmented Hierarchical Analysis and Narration for
                  Financial Data Narration},
  url          = {https://aclanthology.org/2025.findings-emnlp.1405/},
  booktitle    = {{EMNLP} (Findings)},
  pages        = {25761--25785},
  publisher    = {Association for Computational Linguistics},
  year         = {2025}
}

@inproceedings{lewis2020bart,
  author       = {Mike Lewis and
                  Yinhan Liu and
                  Naman Goyal and
                  Marjan Ghazvininejad and
                  Abdelrahman Mohamed and
                  Omer Levy and
                  Veselin Stoyanov and
                  Luke Zettlemoyer},
  title        = {{BART:} Denoising Sequence-to-Sequence Pre-training for Natural Language
                  Generation, Translation, and Comprehension},
  url          = {https://doi.org/10.18653/v1/2020.acl-main.703},
  booktitle    = {{ACL}},
  pages        = {7871--7880},
  publisher    = {Association for Computational Linguistics},
  year         = {2020}
}

@inproceedings{yasunaga2024analogical,
  author       = {Michihiro Yasunaga and
                  Xinyun Chen and
                  Yujia Li and
                  Panupong Pasupat and
                  Jure Leskovec and
                  Percy Liang and
                  Ed H. Chi and
                  Denny Zhou},
  title        = {Large Language Models as Analogical Reasoners},
  url          = {https://openreview.net/forum?id=AgDICX1h50},
  booktitle    = {{ICLR}},
  publisher    = {OpenReview.net},
  year         = {2024}
}

@inproceedings{wiratunga2024cbrrag,
  author       = {Nirmalie Wiratunga and
                  Ramitha Abeyratne and
                  Lasal Jayawardena and
                  Kyle Martin and
                  Stewart Massie and
                  Ikechukwu Nkisi{-}Orji and
                  Ruvan Weerasinghe and
                  Anne Liret and
                  Bruno Fleisch},
  title        = {{CBR-RAG:} Case-Based Reasoning for Retrieval Augmented Generation
                  in LLMs for Legal Question Answering},
  url          = {https://doi.org/10.1007/978-3-031-63646-2\_29},
  booktitle    = {{ICCBR}},
  series       = {Lecture Notes in Computer Science},
  pages        = {445--460},
  publisher    = {Springer},
  year         = {2024}
}

@inproceedings{park2023generative,
  author       = {Joon Sung Park and
                  Joseph C. O'Brien and
                  Carrie Jun Cai and
                  Meredith Ringel Morris and
                  Percy Liang and
                  Michael S. Bernstein},
  title        = {Generative Agents: Interactive Simulacra of Human Behavior},
  url          = {https://doi.org/10.1145/3586183.3606763},
  booktitle    = {{UIST}},
  pages        = {2:1--2:22},
  publisher    = {{ACM}},
  year         = {2023}
}

@inproceedings{zhong2024memorybank,
  author       = {Wanjun Zhong and
                  Lianghong Guo and
                  Qiqi Gao and
                  He Ye and
                  Yanlin Wang},
  title        = {MemoryBank: Enhancing Large Language Models with Long-Term Memory},
  url          = {https://doi.org/10.1609/aaai.v38i17.29946},
  booktitle    = {{AAAI}},
  pages        = {19724--19731},
  publisher    = {{AAAI} Press},
  year         = {2024}
}

@inproceedings{Moosavi2021scigen,
  author       = {Nafise Sadat Moosavi and
                  Andreas R{\"{u}}ckl{\'{e}} and
                  Dan Roth and
                  Iryna Gurevych},
  title        = {SciGen: a Dataset for Reasoning-Aware Text Generation from Scientific
                  Tables},
  url          = {https://datasets-benchmarks-proceedings.neurips.cc/paper/2021/hash/149e9677a5989fd342ae44213df68868-Abstract-round2.html},
  booktitle    = {NeurIPS Datasets and Benchmarks},
  year         = {2021}
}

@inproceedings{shao2023iterative,
  author       = {Zhihong Shao and
                  Yeyun Gong and
                  Yelong Shen and
                  Minlie Huang and
                  Nan Duan and
                  Weizhu Chen},
  title        = {Enhancing Retrieval-Augmented Large Language Models with Iterative
                  Retrieval-Generation Synergy},
  url          = {https://doi.org/10.18653/v1/2023.findings-emnlp.620},
  booktitle    = {{EMNLP} (Findings)},
  series       = {Findings of {ACL}},
  pages        = {9248--9274},
  publisher    = {Association for Computational Linguistics},
  year         = {2023}
}

@inproceedings{trivedi2023interleaving,
  author       = {Harsh Trivedi and
                  Niranjan Balasubramanian and
                  Tushar Khot and
                  Ashish Sabharwal},
  title        = {Interleaving Retrieval with Chain-of-Thought Reasoning for Knowledge-Intensive
                  Multi-Step Questions},
  url          = {https://doi.org/10.18653/v1/2023.acl-long.557},
  booktitle    = {{ACL} {(1)}},
  pages        = {10014--10037},
  publisher    = {Association for Computational Linguistics},
  year         = {2023}
}

@inproceedings{yang2025multimodaldeepresearcher,
  author       = {Zhaorui Yang and
                  Bo Pan and
                  Han Wang and
                  Yiyao Wang and
                  Xingyu Liu and
                  Luoxuan Weng and
                  Yingchaojie Feng and
                  Haozhe Feng and
                  Minfeng Zhu and
                  Bo Zhang and
                  Wei Chen},
  title        = {Multimodal DeepResearcher: Generating Text-Chart Interleaved Reports
                  from Scratch with Agentic Framework},
  url          = {https://doi.org/10.1609/aaai.v40i40.40734},
  booktitle    = {{AAAI}},
  pages        = {34368--34377},
  publisher    = {{AAAI} Press},
  year         = {2026}
}

@inproceedings{press2023measuring,
  author       = {Ofir Press and
                  Muru Zhang and
                  Sewon Min and
                  Ludwig Schmidt and
                  Noah A. Smith and
                  Mike Lewis},
  title        = {Measuring and Narrowing the Compositionality Gap in Language Models},
  url          = {https://doi.org/10.18653/v1/2023.findings-emnlp.378},
  booktitle    = {{EMNLP} (Findings)},
  series       = {Findings of {ACL}},
  pages        = {5687--5711},
  publisher    = {Association for Computational Linguistics},
  year         = {2023}
}

@article{asquith2005information,
title = {Information content of equity analyst reports},
journal = {Journal of Financial Economics},
volume = {75},
number = {2},
pages = {245-282},
year = {2005},
issn = {0304-405X},
doi = {https://doi.org/10.1016/j.jfineco.2004.01.002},
url = {https://www.sciencedirect.com/science/article/pii/S0304405X04001369},
author = {Paul Asquith and Michael B. Mikhail and Andrea S. Au}
}

@article{huang2014evidence,
    author = {Huang, Allen H. and Zang, Amy Y. and Zheng, Rong},
    title = {Evidence on the Information Content of Text in Analyst Reports},
    journal = {The Accounting Review},
    volume = {89},
    number = {6},
    pages = {2151-2180},
    year = {2014},
    month = {11},
    issn = {0001-4826},
    doi = {10.2308/accr-50833},
    url = {https://doi.org/10.2308/accr-50833},
    eprint = {https://publications.aaahq.org/accounting-review/article-pdf/89/6/2151/28471/accr-50833.pdf},
}

@article{huang2018analyst,
author = {Huang, Allen H. and Lehavy, Reuven and Zang, Amy Y. and Zheng, Rong},
title = {Analyst Information Discovery and Interpretation Roles: A Topic Modeling Approach},
journal = {Management Science},
volume = {64},
number = {6},
pages = {2833-2855},
year = {2018},
doi = {10.1287/mnsc.2017.2751},
URL = {https://doi.org/10.1287/mnsc.2017.2751},
eprint = {https://doi.org/10.1287/mnsc.2017.2751}
}

@book{penman2013financial,
  author    = {Penman, Stephen H.},
  title     = {Financial Statement Analysis and Security Valuation},
  edition   = {5th},
  publisher = {McGraw-Hill/Irwin},
  year      = {2013},
  address   = {New York}
}

@book{murphy1999technical,
  title={Technical analysis of the financial markets: A comprehensive guide to trading methods and applications},
  author={Murphy, John J},
  year={1999},
  publisher={Penguin}
}

@article{feinstein1990high,
  title={High agreement but low kappa: I. The problems of two paradoxes},
  author={Feinstein, Alvan R and Cicchetti, Domenic V},
  journal={Journal of clinical epidemiology},
  volume={43},
  number={6},
  pages={543--549},
  year={1990},
  publisher={Elsevier}
}

@inproceedings{ju2023signals,
  author       = {Jia{-}Huei Ju and
                  Yu{-}Shiang Huang and
                  Cheng{-}Wei Lin and
                  Che Lin and
                  Chuan{-}Ju Wang},
  title        = {A Compare-and-contrast Multistage Pipeline for Uncovering Financial
                  Signals in Financial Reports},
  booktitle    = {{ACL} {(1)}},
  pages        = {14307--14321},
  publisher    = {Association for Computational Linguistics},
  year         = {2023}
}

@article{lu2025extraction,
  title={Extraction of characteristic information from financial super-long texts and prediction of corporate violations},
  author={Lu, Hanglin and Zhang, Yongjie and Xu, Jinchang},
  journal={Research in International Business and Finance},
  pages={103079},
  year={2025},
  publisher={Elsevier}
}

@article{mccoy2024embers,
  title={Embers of autoregression show how large language models are shaped by the problem they are trained to solve},
  author={McCoy, R Thomas and Yao, Shunyu and Friedman, Dan and Hardy, Mathew D and Griffiths, Thomas L},
  journal={Proceedings of the National Academy of Sciences},
  volume={121},
  number={41},
  pages={e2322420121},
  year={2024},
  publisher={National Academy of Sciences}
}

@inproceedings{shao2024storm,
  author       = {Yijia Shao and
                  Yucheng Jiang and
                  Theodore A. Kanell and
                  Peter Xu and
                  Omar Khattab and
                  Monica S. Lam},
  title        = {Assisting in Writing Wikipedia-like Articles From Scratch with Large
                  Language Models},
  booktitle    = {{NAACL-HLT}},
  pages        = {6252--6278},
  publisher    = {Association for Computational Linguistics},
  year         = {2024}
}

@inproceedings{hu2022planet,
  author       = {Zhe Hu and
                  Hou Pong Chan and
                  Jiachen Liu and
                  Xinyan Xiao and
                  Hua Wu and
                  Lifu Huang},
  title        = {{PLANET:} Dynamic Content Planning in Autoregressive Transformers
                  for Long-form Text Generation},
  booktitle    = {{ACL} {(1)}},
  pages        = {2288--2305},
  publisher    = {Association for Computational Linguistics},
  year         = {2022}
}
